\documentclass[runningheads]{llncs}
\usepackage{graphicx}
\usepackage{array,booktabs}
\usepackage{url}            

\usepackage{color, colortbl}
\usepackage{arydshln}
\usepackage{tabularray}

\usepackage{wrapfig}
\usepackage{tikz}
\usepackage{comment}
\usepackage{amsmath,amssymb} 
\usepackage{multirow}
\usepackage{mathtools}
\usepackage{paralist}
\usepackage{soul}

\usepackage{xcolor}
\definecolor{backgroundcolor}{HTML}{F2FBe9} 
\definecolor{highlight}{HTML}{B8ED88} 

\usepackage{subcaption}

\usepackage[accsupp]{axessibility}  

\newcommand \ours {HAF~}
\newcommand \CEfinest {Fine Grained Cross-entropy~}
\newcommand \JSD {Soft Hierarchical Consistency~}
\newcommand \dissimilar {Margin Loss~}
\newcommand \weightconst {Geometric Consistency~}

\newcommand \CEfinestsymb {$L_{CE_{fine}}$~}
\newcommand \JSDsymb {$L_{shc}$~}
\newcommand \dissimilarsymb {$L_m$~}
\newcommand \weightconstsymb {$L_{gc}$~}

\input{alldefin}
\begin{document}
\pagestyle{headings}
\mainmatter
\def\ECCVSubNumber{1471}  

\title{Learning Hierarchy Aware Features for Reducing Mistake Severity}

\titlerunning{HAF}
%
\author{Ashima Garg \and 
Depanshu Sani \and 
Saket Anand} 
\authorrunning{A. Garg et al.}
%
\institute{Indraprastha Institute of Information Technology, Delhi, India\\
\email{\{ashimag, depanshus, anands\}@iiitd.ac.in}}
\maketitle

\begin{abstract}
Label hierarchies are often available apriori as part of biological taxonomy or language datasets WordNet. Several works exploit these to learn hierarchy aware features in order to improve the classifier to make semantically meaningful mistakes while maintaining or reducing the overall error. In this paper, we propose a novel approach for learning \textit{Hierarchy Aware Features (HAF)} that leverages classifiers at each level of the hierarchy that are constrained to generate predictions consistent with the label hierarchy. The classifiers are trained by minimizing a Jensen-Shannon Divergence with target soft labels obtained from the fine-grained classifiers. Additionally, we employ a simple geometric loss that constrains the feature space geometry to capture the semantic structure of the label space. \ours is a \textit{training time} approach that improves the mistakes while maintaining top-1 error, thereby, addressing the problem of cross-entropy loss that treats all mistakes as equal. We evaluate \ours on three hierarchical datasets and achieve state-of-the-art results on the iNaturalist-19 and CIFAR-100 datasets. The source code is available at \url{https://github.com/07Agarg/HAF}

\end{abstract}

\section{Introduction}\label{sec:introduction}

Conventional classifiers trained with the cross-entropy loss treat all misclassifications equally. However, certain categories may be more semantically related to each other than to other categories, implying that some classification mistakes may be more \emph{severe} than others. For instance, an autonomous vehicle confusing a {\tt car} for a {\tt truck} is not as severe as mistaking a {\tt pedestrian} for {\tt road}, where the latter mistake could lead to a catastrophe. Similarly, falsely identifying a {\tt pine tree} with an {\tt oak tree} is less severe than identifying it as a {\tt rose}.
Classifiers trained to make mistakes with lower severity could benefit and are often critical in many real-world applications. 

\begin{figure*}[t!]
	\centerline{\includegraphics[width=\textwidth]{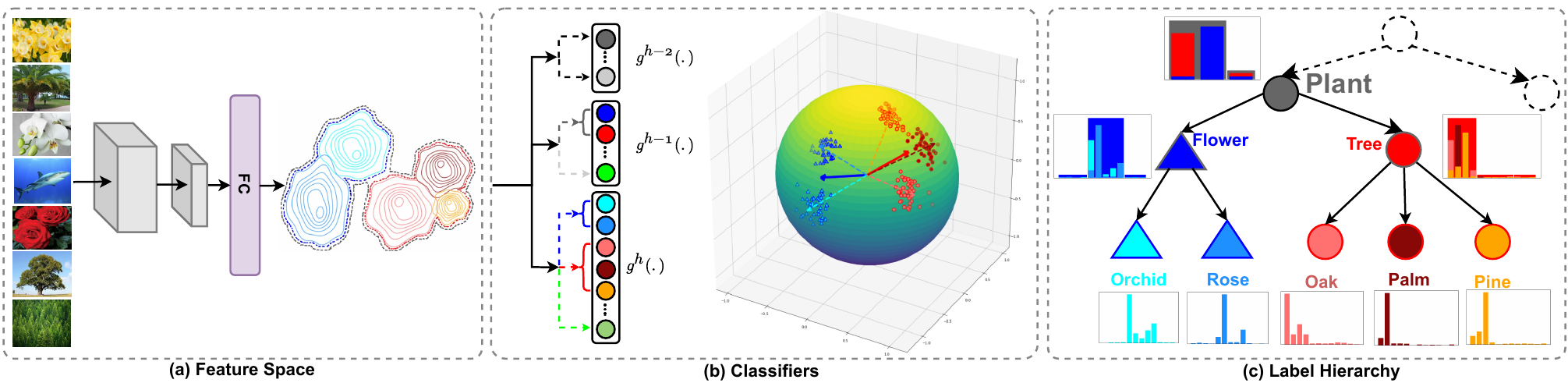}}
	\caption{\textbf{Overview of HAF}. We propose a probabilistic approach using to learn \textit{hierarchy-aware features} that respect the label hierarchy in the feature space and thereby make semantically meaningful mistakes. We train separate classifiers, with a shared feature space, for each level of the label hierarchy. We model the relationship between the fine-grained classes and their respective coarser classes using the label hierarchy and impose consistency constraints on the probability distributions. We further impose simple geometric constraints on the weight vectors of classifiers from different levels to align the weight vectors of fine-grained classes with their corresponding weight vectors of coarser-classes. }
	\label{fig:proposed_approach}
\end{figure*}

The severity of a mistake is typically defined based on some notion of semantic similarity between class labels. For example, a taxonomic hierarchy tree defined over the class labels can express specific semantic relationships between classes through its tree structure, thus enabling an ordering of classes. This ordering was obtained using the lowest common ancestor (LCA) measure in \cite{bertinetto_2020_CVPR,karthik_2021_ICLR}. These hierarchies are often readily available in the class label space as part of language datasets like WordNet \cite{miller_1998_wordnet} or from biological taxonomies, e.g., the one used with the iNaturalist-19 dataset \cite{van_inat_2018_CVPR}.

Bertinetto et al. \cite{bertinetto_2020_CVPR} proposed approaches to reduce the severity of mistakes by employing hierarchy-sensitive adaptations of the cross-entropy loss. They reported reduction in the mistake severity based on the average hierarchical distance of top-$k$ predictions at the cost of an increased top-$1$ error, with the trade-off being controlled by a hyperparameter. A more desirable solution would be the one that reduces the severity of mistakes while maintaining or reducing the overall top-$1$ error. Karthik et al. \cite{karthik_2021_ICLR} highlighted this trade-off and pointed out that the classical approach of Conditional Risk Minimization (CRM) could reduce the mistake severity without a significant change in the top-$1$ error. Moreover, CRM is a test-time intervention that applies post-hoc corrections on the class likelihoods using the LCA measure between classes. Despite its simplicity, the CRM approach is versatile and its effectiveness is remarkable. In Sec. \ref{sec:experiments}, we show that CRM, when combined with other approaches, almost always improves the mistake severity, without a significant impact on the top-$1$ error. 

While CRM improves the quality of prediction errors, being a test-time approach, it does not affect the model. Consequently, the learned representations are inherently inadequate because the cross-entropy loss function ignores all semantic structure in the label space and treats each class independently. 
{\color{red}}{ To overcome this limitation, the hierarchical cross-entropy (HXE) loss was proposed in \cite{bertinetto_2020_CVPR}, which essentially amounts to a weighted combination of the cross-entropy loss applied at different levels of the hierarchy\footnote{{\color{red}} See the supplementary material for a derivation}. 
} Chang et al. \cite{flamingo_2021_Chang} pointed out that training with a coarse class cross-entropy loss deteriorates the accuracy at fine-grained levels. This is likely the reason why both variants proposed in \cite{bertinetto_2020_CVPR}, HXE and the soft-labels loss, result in a trade-off between top-$1$ error and the severity of mistakes. Chang et al. \cite{flamingo_2021_Chang} mitigate this trade-off by disentangling the coarse and the fine-grained features by explicitly partitioning the feature space. This disentangling approach proved to be successful for small hierarchies, however, the feature vector partitioning limits its scalability to larger hierarchies. We argue that for addressing the problem of mistake severity, while maintaining the top-$1$ error, it is important to learn a feature space that captures the structure available in the label space. To this end, we propose learning a \emph{hierarchy-aware feature} (HAF) space that is explicitly trained to inherit the hierarchical structure of the labels. 


We observe that a hierarchy-aware feature space should enable classification at \emph{all} levels of the hierarchy, and simultaneously lead to a lower mistake severity at the finest level. The label hierarchy structure constrains the coarse-level class labels to be a composition of \emph{disjoint} sets of its sub-classes in the hierarchy. We exploit two key properties of the classifiers acting on the feature space to help inherit this compositional structure from the label space. 

{\color{red}}{First, we train a classifier using the fine-grained cross-entropy and use its predictions to obtain target soft labels (Fig. \ref{fig:proposed_approach}) for training the coarse-level auxiliary classifiers. The coarse-level classifiers minimize the Jensen-Shannon divergence (JSD) between their predictions and the target soft labels. This loss avoids the use of hard labels at coarser levels and thus serves as a consistency regularization for the fine-grained classifier, which in turn leads to improved mistake severity without compromising the top-$1$ error. We take this approach to avoid the pitfall highlighted in \cite{flamingo_2021_Chang}, which states that fine-grained features can lead to better coarse-grained predictions, however, explicitly using cross-entropy loss for coarse-level classifiers leads to feature spaces that worsens the performance at a finer granularity.}
Second, we impose geometric consistency constraints on the classifier weight vectors that align sub-classes belonging to the same super-class (Fig. \ref{fig:proposed_approach}(b)). The resulting loss promotes a feature space (Fig. \ref{fig:proposed_approach}(a)) that respects the semantic hierarchy of the label space (Fig. \ref{fig:proposed_approach}(c)). We present further details of the loss terms in Sec. \ref{sec:approach}.   
We summarize our contributions below.
\begin{itemize}
    \item We introduce a novel approach for learning a \textit{hierarchy-aware feature} (\ours) space by inheriting the structure of the label space. We design the loss functions that impose probabilistic and geometric constraints between coarse and fine level classifiers. 
    \item We empirically demonstrate that \ours scales well with large label hierarchies and reduces mistake severity while maintaining the top-$1$ fine-grained error. 
\end{itemize}

\section{Related Work}\label{sec:related_works}
Several works exploit the hierarchical taxonomy of the data for image classification for visual \cite{bertinetto_2020_CVPR,karthik_2021_ICLR,flamingo_2021_Chang} and text \cite{mao_2019_EMNLP} data, multi-label classification tasks \cite{wehrmann_2018_ICML}, image retrieval \cite{barz_2019_WACV,yang_2022_WACV}, object recognition \cite{redmon_2017_yolo_CVPR}, and recently to improve semi-supervised approaches \cite{garg_2022_hiermatch,su_2021_BMVC}. We discuss some of the important works that are closely related with our objective. 

\textbf{Label-embedding methods.} These methods model the class relationships using soft-embeddings. DeViSE \cite{frome_2013_NIPS} maximizes the cosine similarity between the embeddings of an image extracted from a pretrained visual model and the embeddings of label obtained using pretrained word2vec model on Wikipedia. Liu et al. \cite{liu_2020_CVPR} exploit hyperbolic geometry to learn the hierarchical representations. Similar to DeViSE \cite{frome_2013_NIPS}, they minimize the Poincar\'e distance between the Poincar\'e label embeddings \cite{nickel_2017_NIPS} and the image features embeddings. Barz \& Denzler \cite{barz_2019_WACV} map the embeddings onto a unit hypersphere and use LCA to encode the hierarchical distances. Bengio et al. \cite{bengio_2010_NIPS} impose the structure over the classes and fastens learning to embed in low dimensional space to model semantic relationships between classes. Bertinetto et al. \cite{bertinetto_2020_CVPR} proposed Soft-labels that uses the soft-targets encoded with inter-class semantic information based on LCA.

\textbf{Hierarchical-architecture based methods.} Wu et al. \cite{wu_2016_ACM} jointly optimize multi-task loss function wherein cross-entropy loss is applied at each hierarchical level. Recently, Chang et al. \cite{flamingo_2021_Chang} established that jointly optimizing fine-grained with coarse-grained recognition in vanilla framework deteriorates performance on fine-grained classification. The authors proposed architecture for multi-granularity classification with independent level-specific classifiers. Redmon et al. \cite{redmon_2017_yolo_CVPR} proposed a probabilistic model, YOLOv2, for object detection and classification, where softmax is applied at every coarse-category level to address the mutual exclusion of all the classes in conventional softmax classifier. 

\textbf{Hierarchical-loss based methods.} Bertinetto et al. \cite{bertinetto_2020_CVPR} proposed another approaches - hierarchical cross-entropy (HXE). HXE is a probabilistic approach that optimizes a loss function based on conditional probabilities, where predictions for a particular class is conditioned on the parent-class probabilities. Brust \& Denzler et al. \cite{brust_2019_ACPR} proposed a conditional probability classifier for DAGs. Bilal et al. \cite{bilal_2017_IEEE} proposed hierarchical-aware convolutional neural networks by adding branches to the intermediate network pipeline. In \cite{landrieu_2021_BMVC}, authors use prototypical network which uses softmax over distances between the features to the class prototypes, along with a regularization term that encourages the class prototypes to follow the relationship in label hierarchy. Our work is in line with this body of research. We study a different probabilistic model and propose a loss function based on that model. In HAF, we explicitly define class prototypes at every level and take a different approach for arrangement of these prototype vectors. 

\textbf{Cost based methods.} Another line of research is based on assigning different costs depending on the types of misclassification \cite{abe_2004_KDD}. Deng et al. \cite{deng_2010_ECCV} proposed to use \textit{mean classification cost} to make hierarchy-aware predictions by penalizing the mistakes based on the hierarchies. \cite{deng_2012_CVPR,wang_2021_cybernetics} used semantic hierarchy to design cost matrix optimizing accuracy-specificity trade-offs between the level of abstraction of the selected class while selecting the best in specificity. These methods include both internal and leaf nodes in the cost matrix. While Karthik et al. \cite{karthik_2021_ICLR} study conditional risk minimization (CRM) on similar lines to \cite{deng_2010_ECCV}, an inference-time approach that weighs the predictions based on the cost matrix defined using LCA distances among the leaf nodes. \ours also fits in this framework. However, unlike CRM \cite{karthik_2021_ICLR}, \ours is a training-time approach to learn feature embeddings such that they are hierarchically meaningful. 





\section{HAF: Proposed Approach}\label{sec:approach}

Consider a label hierarchy tree with $H+1$ levels, where the root is at level-0, and $h\in[1,\ldots,H]$ denote the hierarchical level with $h\!=\!1$ and $h\!=\!H$ the coarsest and finest levels respectively. We ignore the root node for our purposes as it denotes the universal super-set containing all classes. Let $\cal{X}$ = $\{ \bx_i, y_i^h \vert i=1,\ldots,N \}$ be the set of $N$ images and their respective ground-truth labels at level $h$. We denote the common feature extractor $f_{\phi}(\cdot)$, which is implemented using some backbone neural network and is parameterized by $\phi$. As illustrated in Fig. \ref{fig:proposed_approach}, we use classifiers at each level of the hierarchy in training \ours and denote the level-$h$ classifier as $g^h(\cdot)$ parameterized by the weight matrix $\bW^h$. The resulting prediction probabilities are denoted by $ p^h(\widehat{y}_i^h|\bx_i;\bW^h) = g^h(f_{\phi}(\bx_i))$, where $\widehat{y}_i^h$ is the label predicted for $\bx_i$ by $g^h(\cdot)$ and can take class labels from the set of classes at level-$h$ as $\calC^{h}=\left\{ \bigcup_{i=1}^{|A|} A_i, \bigcup_{i=1}^{|B|} B_i, \bigcup_{i=1}^{|C|} C_i, \ldots\right\}$, where we define the set of classes at level-$(h-1)$ as $\calC^{h-1}=\{A,B,C,\ldots\}$. With a slight abuse of notation, here we use $A$ to denote a super-class label at level-$(h-1)$ and the set of its sub-classes $\{A_1,A_2,\ldots\}$ at level-$h$. 



\subsection{\CEfinest  ($L_{CE_{fine}}$)} 

We use the ground truth labels only at the finest level of the hierarchy and apply the cross-entropy loss to train the level-$H$ classifier, i.e.,  ${g}^H(\cdot)$. The fine-grained cross-entropy loss for a sample is given by
\begin{gather}
L_{CE_{fine}} = - \sum_{c\in\calC^H} \mathbf{1}\!\left[y_i^H\!=\!c\right]\log \left(p^H(\widehat{y}_i^H\!=\!c|\bx_i;\bW^H)\right)
\label{eqn:cross-entropy}
\end{gather}
where $\mathbf{1}[\cdot]$ serves as an indicator function and takes a value of one when the argument is true, else zero.

\subsection{\JSD ($L_{shc}$)}
\begin{wrapfigure}[24]{r}{0.44\textwidth}
\centering{
    \includegraphics[width=0.42\textwidth]{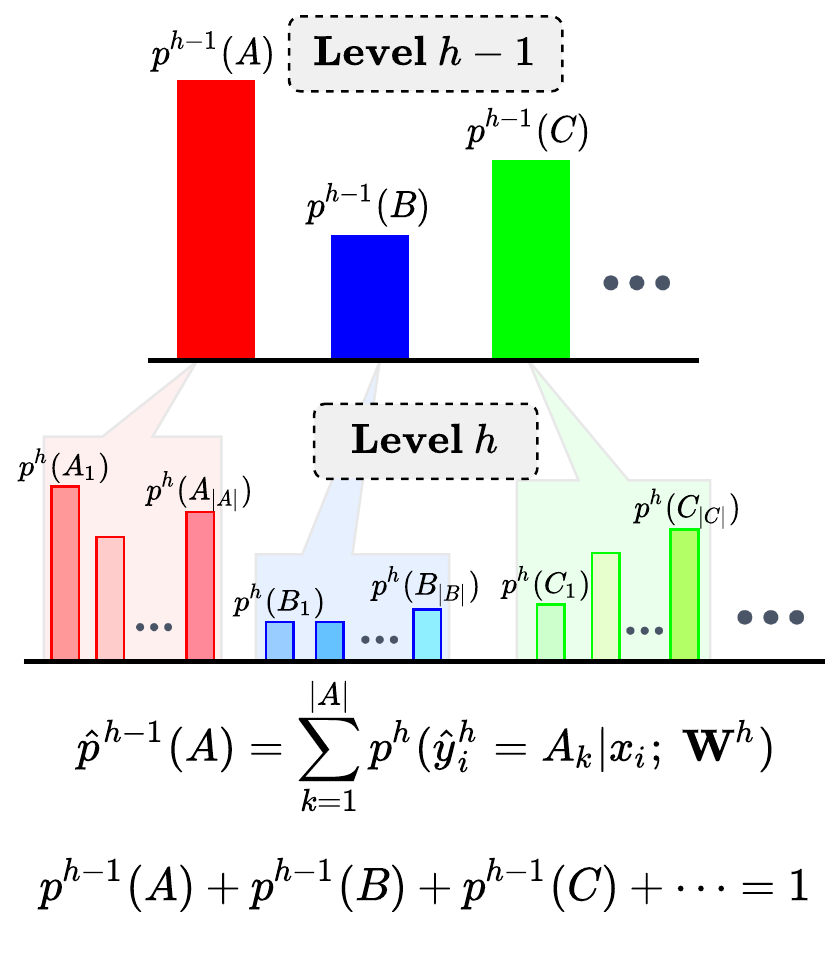}
    }
  \caption{Constructing the soft labels for training the coarse-level classifiers. The super-class target probability is the sum of its sub-classes' predicted probability. The colors indicate the class relationships across levels $h-1$ and $h$.}
  \label{fig:notations}
\end{wrapfigure}

For making better mistakes, we want the classifiers at all levels to use the same feature space and yet make predictions consistent with the label hierarchy. While it is natural to use the cross-entropy loss for training the classifiers at all levels, as noted in \cite{flamingo_2021_Chang} and observed during our initial experiments, this choice of loss compromises the fine-grained accuracy. Instead, we enforce the consistency across classifiers at different levels by using soft labels and a symmetric entropy-based loss function. We minimize the Jensen-Shannon Divergence (JSD) \cite{fuglede_2004_jensen} between the predictions of a coarse classifier $g^{h-1}(\cdot)$ and the soft labels obtained from the next fine-level classifier $g^h(\cdot)$. As defined above, for a given class label $A\in\calC^{h-1}$, let $P[\widehat{y}^{h-1}\!=\!A|\bx_i]$ denote the probability of the sample $\bx_i$ belonging to the class $A$, which is computed as 
\begin{gather}
\label{eqn:total_prob}
P\left[\widehat{y}_i^{h-1}\!=\!A|\bx_i\right] = \sum_{k=1}^{|A|} p^h(\widehat{y}_i^h=A_k|\bx_i;\bW^{h})
\end{gather}
The probabilities $P[c], \forall c\in\calC^{h-1}$ are concatenated together to construct the probability vector $\widehat{p}^{h-1}(\widehat{y}_i^{h-1}|\bx_i)$, which is used as the soft label for $\bx_i$. This soft label generation process is illustrated in Fig. \ref{fig:notations}.

The JSD is minimized between the soft labels and the predictions from the classifier $g^h(\cdot)$. For convenience, we use $p_i^h$ to refer to $ p^h(\hat{y_i}^h|x_i;\bW^h) $ and similarly $\widehat{p}_i^h$ for the corresponding soft label. The JSD based total \JSD is computed by summing the pairwise losses across the levels
\begin{gather}
\text{L}_{shc} = \sum_{h=1}^{H-1}JS^h\left(p_i^h||\widehat{p}_i^h\right) = \frac{1}{2} \sum_{h=1}^{H-1} (\text{KL}(p_i^h || m) + \text{KL}(\widehat{p}_i^h || m))
\label{eqn:shc_loss}
\end{gather}
where $m$ = $\frac{1}{2}(p_i^h + \widehat{p}_i^h)$ and KL$(\cdot||\cdot)$ refers to Kullback-Leibler divergence. 

It is important to highlight the key difference between the soft labels generated above and those defined in \cite{bertinetto_2020_CVPR}. The latter are designed using the LCA-based distance between classes, whereas our choice of soft labels can be interpreted as a \emph{learned} label-smoothing that better regularizes the coarse-level classifiers. Yuan et al. \cite{Yuan_2020_CVPR} make a similar argument about label smoothing in the context of knowledge distillation. The use of a symmetric loss like in eqn. (\ref{eqn:shc_loss}) further enables the classifiers at both levels to jointly drive the feature space learning. This behavior of the coarse classifiers improving the performance of the finer-level classifiers is analogous to the Reversed Knowledge Distillation (Re-KD) setting as presented in \cite{Yuan_2020_CVPR}, where the authors showed that a student ($g^{h-1}(\cdot)$) is capable of improving the performance of the teacher ($g^h(\cdot)$).  

\subsection{\dissimilar ($L_m$)}
While $\text{L}_{shc}$ improves the mistake severity (as we show in Sec. \ref{sec:analysis}) successfully by virtue of better regularization, it does not directly encourage discrimination between coarse-level classes. Therefore, we use a pairwise margin-based loss to promote a more discriminative feature space. We use this loss over coarser levels $h \in \calH$ where $\calH$ is [$k$, $H-1$] and $k$ ranges from [$1$ , $H-1$]. For a given batch of samples, we create pairs of samples that have dissimilar labels at a level $h$, i.e., $\calB^h = \{(i,j) | y^h_i\neq y^h_j\}$. Then we compute the margin loss over the batch as
\begin{gather}
    L_{m} = \sum_{h\in\calH} \sum_{(i,j)\in\calB^h}\max\left(0, m - \text{JS}^h(p^h_i || p^h_j)\right)
    \label{eqn:margin_loss}
\end{gather}
where $p^h_i$ is the softmax probability generated by $g^h(f_\phi(\bx_i))$ and $m$ is the margin. The margin loss is only applied to the coarser levels of the hierarchy, as the cross-entropy loss of (\ref{eqn:cross-entropy}) is sufficient for fine-grained discrimination. 

\subsection{\weightconst ($L_{gc}$)}
\ours uses classifiers at all the levels of hierarchy. In a hierarchy-aware feature space, the weight vectors of the coarse class and its fine-grained classes should be correlated. The losses introduced in the previous subsections impose probabilistic consistency across the classifier predictions, and only indirectly affect the feature space geometry. In order to better orient the feature space to inherit the label space hierarchy, we use a geometric consistency loss. As before, let $A\in\calC^{h-1}$ be a given super-class and its sub-classes be $A_k\in\calC^h,~k=1,\ldots,|A|$. Let the weight vector corresponding to the super-class $A$ be $\bw^{h-1}_A$ and similarly the weight vectors corresponding to the sub-classes be $\bw^{h}_{A_k}$. Note that the classifier $g^{h-1}(\cdot)$ is defined by the weight matrix $\bW^{h-1}$, which is obtained by stacking the weight vectors $\bw_c,~c\in\calC^{h-1}$. We further constrain each weight vector to be unit norm $||\bw^h_c||_2=1, \forall c,h $, across all classifiers. For the super-class $A\in\calC^{h-1}$, we define the target weight vector as $\widehat{\bw}_A^{h-1} = \widetilde{\bw}^{h-1}_A/||\widetilde{\bw}_A^{h-1}||_2$, where $\widetilde{\bw}_A^{h-1}=\sum_{k=1}^{|A|}\bw_{A_k}^h$. Thus, the \weightconst loss to be minimized is 
\begin{gather}
    L_{gc} = \sum_{h=1}^{H-1} \sum_{c\in\calC^h}\left(1 - \cos\left(\bw_c^{h},\widehat{\bw}_c^h\right)\right)
    \label{eqn:gc_loss}
\end{gather}
where $\cos\left(\bw_c^{h},\widehat{\bw}_c^h\right)$ refers to the cosine similarity between the weight vectors $\bw_c^{h}$ and $\widehat{\bw}_c^h$.

Finally, the total loss is given by $L_{total} =$  \CEfinestsymb + \JSDsymb + \dissimilarsymb + \weightconstsymb .

\section{Experiments and Results}\label{sec:experiments}
\subsection{Experimental Setup}

\noindent \textbf{Datasets.}
We present the evaluation of \ours approach on the CIFAR-100 \cite{krizhevsky_2009_CIFAR100}, iNaturalist-19 \cite{van_inat_2018_CVPR} and tieredImageNet-H \cite{ren_tiered_18_ICLR} datasets. 
We follow the hierarchical taxonomy as is in \cite{landrieu_2021_BMVC} for CIFAR-100, and \cite{bertinetto_2020_CVPR} for iNaturalist-19 and tieredImageNet-H. 
In all the three datasets, Level-0 has only one node, i.e., the root node. Therefore, we only consider the bottom $H$ hierarchical levels. Similar to \cite{bertinetto_2020_CVPR}, we compute the distance between any two nodes by finding the minimum distance between the node and their Lowest Common Ancestor (LCA). Table \ref{tab:dataset_statistics} summarizes the dataset statistics. 
\begin{table}
	\begin{center}
	\begin{tabular}{ l ccccc }
		\hline
		 & ~~Train~~ & ~~Val~~ & ~~Test~~ & ~~\#Classes~~ & ~~\#Levels~~ \\ \hline
		CIFAR-100~~ & ~~45,000~~ & ~~5,000~~ & ~~10,000~~ & ~~100~~ & ~~6~~ \\ 
		iNaturalist-19~~ & ~~187,385~~ & ~~40,121~~ & ~~40,737~~ & ~~1010~~ & ~~8~~ \\ 
		tieredImageNet-H~~ & ~~425,600~~ & ~~15,200~~ & ~~15,200~~ & ~~608~~ & ~~13~~ \\ \hline
	\end{tabular}
	\end{center} 
	\caption{Statistics of the datasets.} 
	\label{tab:dataset_statistics}	
\end{table}

\noindent \textbf{Baselines.} We directly compare \ours with the baseline cross-entropy, Barz \& Denzler's \cite{barz_2019_WACV}, YOLO-v2 \cite{redmon_2017_yolo_CVPR}, both approaches of Bertinetto et al's \cite{bertinetto_2020_CVPR} work - soft-labels and HXE, and the recently proposed CRM-based method from \cite{karthik_2021_ICLR}. We also compare with recently proposed Chang et al.'s \cite{flamingo_2021_Chang} multi-task framework for classification with different granularities. For fair comparisons, we re-run all the experiments with the same codebase under the new best hyperparameter settings for all the methods and report mean and standard deviation of each experiment averaged over three-different seeds. 
\newline
\noindent \textbf{Evaluation Metrics.}
We use the same evaluation metrics as Bertinetto et al.  \cite{bertinetto_2020_CVPR}; Karthik et al. \cite{karthik_2021_ICLR}. We report the following three metrics: 
\begin{inparaenum}[i)]
\item top-$1$ error, \item average mistakes severity, i.e., average LCA-based distance between the  ground-truth and predicted class label for \textit{only} incorrectly classified samples, and \item average hierarchical distance @k, i.e., average distance from the LCA of ground-truth label and $k$ most likely predictions for \textit{all} the samples. 
\end{inparaenum} 

\subsection{Training Configurations}
We adopt the Wideresnet-28-2 \cite{zagoruyko_2016_BMVC} backbone for evaluation on the CIFAR-100 dataset. For the iNaturalist-19 and tieredImageNet-H datasets, we use the ImageNet pretrained ResNet-18 \cite{he_2016_CVPR} backbone with an additional fully-connected (FC) layer of 600 hidden units. Chang et al. \cite{flamingo_2021_Chang} only employ this fully connected layer for facilitating disentanglement, however, we use this additional layer as part of the backbone for consistency across all the methods. Classifiers for each hierarchical level follow this layer. We train all the models with a batch size of 256. We use a fixed margin $m$ of 3.0 across all the datasets defined in Eq (\ref{eqn:margin_loss}) and create a total of $256$ dissimilar pairs from a batch of data.   
For CIFAR-100, we employ \texttt{RandomPadandCrop(32)} and \texttt{RandomFlip()} for augmentation. For iNaturalist-19 and tieredImagenet-H, we use \texttt{RandomHorizontalFlip()} followed by \texttt{RandomResizedCrop()} as carried out in \cite{bertinetto_2020_CVPR}. 


We find the training strategy (learning rate and optimizer) of Chang et al. \cite{flamingo_2021_Chang} to give optimal results on both CIFAR-100 and iNaturalist-19 datasets on the baseline cross-entropy. This training strategy with the SGD optimizer boosts the performance of cross-entropy on iNaturalist-19 as opposed to the ones reported using Adam optimizer in \cite{bertinetto_2020_CVPR}. 
We obtain the best results for CIFAR-100 and iNaturalist-19 using the SGD optimizer on all methods, except for soft-labels and HXE where Adam \cite{j_2018_Adam} performs the best. For the methods trained with SGD, we set different learning rates for the backbone network and the FC layer as $0.01$ and $0.1$ respectively, following \cite{flamingo_2021_Chang}. For training with soft-labels and HXE with Adam optimizer, using a hyperparameter sweep we find that the model performs the best with learning rate as $1e-3$ and $1e-4$ for CIFAR-100 and iNaturalist-19 respectively. We train all the models on tieredImageNet-H for 120 epochs with a learning rate of $1e-5$. Unlike other datasets, we employ the Adam optimizer for tieredImageNet-H as it performed better than the SGD optimizer. 


\subsection{Results}
Tables \ref{tab:cifar-100}, \ref{tab:inat} and \ref{tab:tiereimagenet} present the comparisons of our proposed technique with the baselines on CIFAR-100, iNaturalist-19, and  tieredImageNet-H respectively. Karthik et al. \cite{karthik_2021_ICLR} apply the CRM technique on the baseline cross-entropy. Since CRM is a test-time approach that reweighs the probability distribution of samples obtained from any trained model, it can be applied to all other approaches. Therefore, in each of the Tables \ref{tab:cifar-100}-\ref{tab:tiereimagenet}, we group the results to report evaluation metrics with and without using CRM at test-time. We re-emphasize that the goal of the problem is to improve the hierarchical metrics by maintaining or improving the top-$1$ error. Towards this goal, in each table, we highlight the competitive methods (rows) on the top-$1$ error with \colorbox{backgroundcolor}{lightgreen}. Among these competitive methods, we highlight the best-performing entries for each metric with \colorbox{highlight}{green}. On CIFAR-100 (Table \ref{tab:cifar-100}), baseline cross-entropy, Chang et al.\cite{flamingo_2021_Chang} and \ours and their counterparts using CRM are competitive methods on top-$1$ error. However, \ours and \ours + CRM outperforms all other hierarchical metrics without compromising top-$1$ error. We observe similar trends on iNaturalist-19 (Table \ref{tab:inat}), where \ours, and \ours + CRM are the only competitive \textit{training method} to cross-entropy, which maintain the top-$1$ error and yet improve the hierarchical metrics. On tieredImageNet-H (Table \ref{tab:tiereimagenet}), baseline cross-entropy, HXE $\alpha=0.1$, Soft-labels $\beta=30$, and \ours are competitive for both, top-$1$ error and hierarchical metrics. However, \ours is the best performing method on hier dist@20. 

\begin{table*}[h!]
\begin{center}
\resizebox{0.9\textwidth}{!}{
\begin{tabular}{ l   ccccc } 
\hline
\addlinespace[0.2cm]
\multicolumn{1}{c}{\multirow{2}{*}{Method}} & Top-$1$ Error($\downarrow$) & Mistakes severity($\downarrow$) & Hier dist@1($\downarrow$) & Hier dist@5($\downarrow$) & Hier dist@20($\downarrow$) \\
\cline{2-6} 
       & \multicolumn{5}{c}{Without CRM}                          \\
\addlinespace[0.1cm]
\hline
\rowcolor{backgroundcolor}Cross-Entropy & 22.27 $\pm$ 0.001 & 2.35 $\pm$ 0.024 & 0.52 $\pm$ 0.003 & 2.24 $\pm$ 0.007 & 3.17 $\pm$ 0.007\\
Barz \& Denzler & 31.69 $\pm$ 0.004 & 2.36 $\pm$ 0.025 & 0.75 $\pm$ 0.012 & 1.25 $\pm$ 0.364 & 2.49 $\pm$ 0.004 \\
YOLO-v2 \cite{redmon_2017_yolo_CVPR} & 32.03 $\pm$ 0.006 & 3.72 $\pm$ 0.022 & 1.19 $\pm$ 0.019 & 2.85 $\pm$ 0.010 & 3.39 $\pm$ 0.0109 \\
HXE $\alpha$=0.1 \cite{bertinetto_2020_CVPR} & 28.41 $\pm$ 0.003 & 2.43 $\pm$ 0.004 & 0.69 $\pm$ 0.008 & 2.08 $\pm$ 0.008 & 3.02 $\pm$ 0.012 \\
HXE $\alpha$=0.6 \cite{bertinetto_2020_CVPR} & 30.42 $\pm$ 0.003 & 2.29 $\pm$ 0.008 & 0.7 $\pm$ 0.008 & 1.76 $\pm$ 0.007 & 2.79 $\pm$ 0.008 \\
Soft-labels $\beta=30$ \cite{bertinetto_2020_CVPR} & 26.99 $\pm$ 0.003 & 2.38 $\pm$ 0.004 & 0.64 $\pm$ 0.008 & 1.39 $\pm$ 0.027 & 2.79 $\pm$ 0.005 \\
Soft-labels $\beta=4$ \cite{bertinetto_2020_CVPR} & 32.15 $\pm$ 0.008 & 2.21 $\pm$ 0.037 & 0.71 $\pm$ 0.024 & 1.23 $\pm$ 0.018 & 2.23 $\pm$ 0.008 \\
\rowcolor{backgroundcolor}Chang et al. \cite{flamingo_2021_Chang} & \cellcolor{highlight}21.94 $\pm$ 0.002 & 2.32 $\pm$ 0.005 & 0.51 $\pm$ 0.005 & 2.06 $\pm$ 0.018 & 3.08 $\pm$ 0.007 \\
\rowcolor{backgroundcolor}\ours & 22.27 $\pm$ 0.001 & \cellcolor{highlight} 2.24 $\pm$ 0.014 & \cellcolor{highlight}0.50 $\pm$ 0.003 & \cellcolor{highlight}1.41 $\pm$ 0.007 & \cellcolor{highlight}2.64 $\pm$ 0.002 \\
\addlinespace[0.1cm]
\hline
       & \multicolumn{5}{c}{With CRM}                          \\
       \cline{2-6} 
\addlinespace[0.1cm]
\rowcolor{backgroundcolor}Cross-Entropy \cite{karthik_2021_ICLR} & 22.23 $\pm$ 0.001 & 2.31 $\pm$ 0.033 & 0.51 $\pm$ 0.006 & 1.11 $\pm$ 0.006 & 2.18 $\pm$ 0.002 \\
YOLO-v2 & 32.01 $\pm$ 0.006 & 3.72 $\pm$ 0.020 & 1.19 $\pm$ 0.021 & 3.17 $\pm$ 0.003 & 3.64 $\pm$ 0.004 \\
HXE ($\alpha$=0.1) & 28.41 $\pm$ 0.003 & 2.42 $\pm$ 0.005 & 0.69 $\pm$ 0.007 & 1.24 $\pm$ 0.005 & 2.24 $\pm$ 0.005 \\
HXE ($\alpha$=0.6) & 30.46 $\pm$ 0.003 & 2.28 $\pm$ 0.009 & 0.69 $\pm$ 0.009 & 1.22 $\pm$ 0.007 & 2.22 $\pm$ 0.004 \\
Soft-labels ($\beta=30$) & 27.17 $\pm$ 0.004 & 2.36 $\pm$ 0.001 & 0.64 $\pm$ 0.008 & 1.20 $\pm$ 0.005 & 2.22 $\pm$ 0.003 \\
Soft-labels ($\beta=4$) & 32.73 $\pm$ 0.007 & 2.21 $\pm$ 0.023 & 0.72 $\pm$ 0.017 & 1.23 $\pm$ 0.011 & 2.23 $\pm$ 0.006 \\
\rowcolor{backgroundcolor}Chang et al. \cite{flamingo_2021_Chang} & \cellcolor{highlight}21.92 $\pm$ 0.001 & 2.27 $\pm$ 0.009 & \cellcolor{highlight}0.50 $\pm$ 0.003 & \cellcolor{highlight}1.10 $\pm$ 0.002 & 2.18 $\pm$ 0.002 \\
\rowcolor{backgroundcolor}\ours & 22.31 $\pm$ 0.001 & \cellcolor{highlight}2.23 $\pm$ 0.018 & \cellcolor{highlight}0.50 $\pm$ 0.003 & \cellcolor{highlight}1.10 $\pm$ 0.003 & \cellcolor{highlight} 2.17 $\pm$ 0.003 \\
\addlinespace[0.1cm]
\hline
\end{tabular}
}
\end{center}
\caption{Results comparing top-$1$ error(\%) and hierarchical metrics on the test set of \textit{CIFAR-100}. Results in the \textit{Top} block are reported without using CRM \cite{karthik_2021_ICLR} technique and \textit{Bottom} block are reported using CRM. Rows highlighted with \colorbox{backgroundcolor}{lightgreen} are competitive methods in top-$1$ error (\%). Of these competitive methods, we highlight the best performing entries for each metric with \colorbox{highlight}{green}.}
\label{tab:cifar-100}
\end{table*}

\begin{table*}[h!]
\begin{center}
\resizebox{0.9\textwidth}{!}{
\begin{tabular}{ l  ccccc } 
\hline
\addlinespace[0.2cm]
\multicolumn{1}{c}{\multirow{2}{*}{Method}} & Top-$1$ Error($\downarrow$) & Mistakes severity($\downarrow$) & Hier dist@1($\downarrow$) & Hier dist@5($\downarrow$) & Hier dist@20($\downarrow$) \\
\cline{2-6} 
       & \multicolumn{5}{c}{Without CRM}                          \\
\addlinespace[0.1cm]
\hline
\rowcolor{backgroundcolor}Cross-Entropy & 36.44 $\pm$ 0.061 & 2.39 $\pm$ 0.007 & 0.87 $\pm$ 0.004 & 1.97 $\pm$ 0.002 & 3.25 $\pm$ 0.002 \\
Barz \& Denzler \cite{barz_2019_WACV} & 62.63 $\pm$ 0.278 & 1.99 $\pm$ 0.008 & 1.24 $\pm$ 0.005 & 1.49 $\pm$ 0.005 & 1.97 $\pm$ 0.005 \\
YOLO-v2 \cite{redmon_2017_yolo_CVPR} & 44.37 $\pm$ 0.106 & 2.42 $\pm$ 0.003 & 1.08 $\pm$ 0.004 & 1.90 $\pm$ 0.003 & 2.87 $\pm$ 0.010 \\
HXE $\alpha$=0.1 \cite{bertinetto_2020_CVPR} & 41.48 $\pm$ 0.204 & 2.41 $\pm$ 0.009 & 1.00 $\pm$ 0.006 & 1.77 $\pm$ 0.011 & 2.69 $\pm$ 0.021 \\
HXE $\alpha$=0.6 \cite{bertinetto_2020_CVPR} & 45.45 $\pm$ 0.014 & 2.24 $\pm$ 0.006 & 1.02 $\pm$ 0.003 & 1.70 $\pm$ 0.005 & 2.55 $\pm$ 0.005\\
Soft-labels $\beta=30$ \cite{bertinetto_2020_CVPR} & 41.67 $\pm$ 0.134 & 2.32 $\pm$ 0.010 & 0.97 $\pm$ 0.006 & 1.50 $\pm$ 0.006 & 2.23 $\pm$ 0.005 \\
Soft-labels $\beta=4$ \cite{bertinetto_2020_CVPR} & 74.70 $\pm$ 0.212 & 1.82 $\pm$ 0.005 & 1.36 $\pm$ 0.004 & 1.49 $\pm$ 0.003 & 1.96 $\pm$ 0.004\\
Chang et al. \cite{flamingo_2021_Chang} & 37.23 $\pm$ 0.175 & 2.28 $\pm$ 0.006 & 0.85 $\pm$ 0.004 & 1.75 $\pm$ 0.005 & 3.02 $\pm$ 0.008 \\
\rowcolor{backgroundcolor}\ours & \cellcolor{highlight}36.4 $\pm$ 0.092 & \cellcolor{highlight}2.28 $\pm$ 0.012 & \cellcolor{highlight}0.83 $\pm$ 0.002 & \cellcolor{highlight}1.62 $\pm$ 0.002 & \cellcolor{highlight}2.55 $\pm$ 0.003 \\
\addlinespace[0.1cm]
\hline
       & \multicolumn{5}{c}{With CRM}                          \\
       \cline{2-6} 
\addlinespace[0.1cm]
\rowcolor{backgroundcolor}Cross-Entropy \cite{karthik_2021_ICLR} & 36.51 $\pm$ 0.083 & 2.33 $\pm$ 0.001 & 0.85 $\pm$ 0.002 & 1.32 $\pm$ 0.001 & 1.86 $\pm$ 0.002 \\
YOLO-v2 & 45.17 $\pm$ 0.046 & 2.43 $\pm$ 0.001 & 1.10 $\pm$ 0.001 & 1.50 $\pm$ 0.001 & 1.99 $\pm$ 0.002 \\
HXE $\alpha$=0.1 & 41.47 $\pm$ 0.220 & 2.38 $\pm$ 0.011 & 0.99 $\pm$ 0.008 & 1.41 $\pm$ 0.006 & 1.93 $\pm$ 0.005 \\
HXE $\alpha$=0.6 & 45.60 $\pm$ 0.017 & 2.21 $\pm$ 0.008 & 1.01 $\pm$ 0.003 & 1.40 $\pm$ 0.004 & 1.40 $\pm$ 0.004 \\
Soft-labels $\beta=30$ & 41.99 $\pm$ 0.126 & 2.31 $\pm$ 0.009 & 0.97 $\pm$ 0.007 & 1.40 $\pm$ 0.005 & 1.91 $\pm$ 0.005 \\
Soft-labels $\beta=4$ & 77.34 $\pm$ 0.262 & 2.06 $\pm$ 0.012 & 1.60 $\pm$ 0.007 & 1.72 $\pm$ 0.008 & 2.14 $\pm$ 0.007 \\
Chang et al. \cite{flamingo_2021_Chang} & 37.31 $\pm$ 0.145 & 2.24 $\pm$ 0.008 & 0.84 $\pm$ 0.002 & 1.30 $\pm$ 0.002 & 1.84 $\pm$ 0.002 \\
\rowcolor{backgroundcolor}\ours &  \cellcolor{highlight}36.48 $\pm$ 0.095 &  \cellcolor{highlight}2.25 $\pm$ 0.012 &  \cellcolor{highlight}0.82 $\pm$ 0.003 & \cellcolor{highlight}1.29 $\pm$ 0.004 &  \cellcolor{highlight}1.84 $\pm$ 0.002 \\
\addlinespace[0.1cm]
\hline
\end{tabular}
}
\end{center}
\caption{Results comparing top-$1$ error(\%) and hierarchical metrics on the test set of \textit{iNaturalist-19}. Results in the \textit{Top} block are reported without using CRM \cite{karthik_2021_ICLR} technique and \textit{Bottom} block are reported using CRM. Rows highlighted with \colorbox{backgroundcolor}{lightgreen} are competitive methods in top-$1$ error (\%). Of these competitive methods, we highlight the best performing entries for each metric with \colorbox{highlight}{green}.} 
\label{tab:inat}
\end{table*}

\begin{table*}[!t]
\begin{center}
\resizebox{0.9\textwidth}{!}{
\begin{tabular}{ l  ccccc } 
\hline
\addlinespace[0.2cm]
\multicolumn{1}{c}{\multirow{2}{*}{Method}} & Top-$1$ error($\downarrow$) & Mistakes severity($\downarrow$) & Hier dist@1($\downarrow$) & Hier dist@5($\downarrow$) & Hier dist@20($\downarrow$) \\
\cline{2-6} 
       & \multicolumn{5}{c}{Without CRM}                          \\
\addlinespace[0.1cm]
\hline
\rowcolor{backgroundcolor}Cross-Entropy & 30.60 $\pm$ 0.030 & 7.05 $\pm$ 0.010 & 2.16 $\pm$ 0.006 & 5.67 $\pm$ 0.003 & 7.17 $\pm$ 0.003 \\
Barz \& Denzler \cite{barz_2019_WACV} & 39.73 $\pm$ 0.240 & 6.80 $\pm$ 0.019 & 2.70 $\pm$ 0.022 & 5.48 $\pm$ 0.271 & 6.21 $\pm$ 0.005 \\
YOLO-v2 \cite{redmon_2017_yolo_CVPR} & 33.37 $\pm$ 0.082 & 7.02 $\pm$ 0.004 & 2.34 $\pm$ 0.016 & 5.85 $\pm$ 0.011 & 7.43 $\pm$ 0.016 \\
DeViSE \cite{frome_2013_NIPS} & 36.75 $\pm$ 0.090 & 6.87 $\pm$ 0.017 & 2.52 $\pm$ 0.009 & 5.57 $\pm$ 0.005 & 6.98 $\pm$ 0.005 \\
\rowcolor{backgroundcolor}HXE $\alpha$=0.1 \cite{bertinetto_2020_CVPR} & 30.72 $\pm$ 0.036 & \cellcolor{highlight}7.00 $\pm$ 0.019 & 2.15 $\pm$ 0.005 & \cellcolor{highlight}5.62 $\pm$ 0.008 & 7.08 $\pm$ 0.015\\
HXE $\alpha$=0.6 \cite{bertinetto_2020_CVPR} & 34.50 $\pm$ 0.007 & 6.73 $\pm$ 0.014 & 2.32 $\pm$ 0.003 & 5.48 $\pm$ 0.001 & 6.78 $\pm$ 0.003\\
\rowcolor{backgroundcolor}Soft-labels $\beta=30$ \cite{bertinetto_2020_CVPR} & 30.53 $\pm$ 0.194 & 7.05 $\pm$ 0.009 & 2.15 $\pm$ 0.013 & 5.66 $\pm$ 0.002 & 7.14 $\pm$ 0.008\\
Soft-labels $\beta=4$ \cite{bertinetto_2020_CVPR} & 38.99 $\pm$ 0.105 & 6.60 $\pm$ 0.024 & 2.57 $\pm$ 0.004 & 5.13 $\pm$ 0.002 & 6.21 $\pm$ 0.001 \\
Chang et al. \cite{flamingo_2021_Chang} & 33.46 $\pm$ 0.026 & 6.99 $\pm$ 0.010 & 2.34 $\pm$ 0.006 & 5.75 $\pm$ 0.005 & 7.34 $\pm$ 0.010 \\
\rowcolor{backgroundcolor}\ours & \cellcolor{highlight}30.50 $\pm$ 0.010 & 7.03 $\pm$ 0.024 & \cellcolor{highlight}2.14 $\pm$ 0.008 & \cellcolor{highlight}5.62 $\pm$ 0.011 & \cellcolor{highlight}6.99 $\pm$ 0.009 \\
\addlinespace[0.1cm]
\hline
       & \multicolumn{5}{c}{With CRM}                          \\
       \cline{2-6} 
\addlinespace[0.1cm]
\rowcolor{backgroundcolor}Cross-Entropy \cite{karthik_2021_ICLR} & 30.67 $\pm$ 0.020 & 6.99 $\pm$ 0.007 & \cellcolor{highlight}2.14 $\pm$ 0.006 & 4.95 $\pm$ 0.002 & \cellcolor{highlight}6.11 $\pm$ 0.001 \\
YOLO-v2 & 33.98 $\pm$ 0.099 & 6.99 $\pm$ 0.011 & 2.38 $\pm$ 0.012 & 5.05 $\pm$ 0.001 & 6.17 $\pm$ 0.001 \\
\rowcolor{backgroundcolor}HXE $\alpha$=0.1 & 30.80 $\pm$ 0.079 & \cellcolor{highlight}6.95 $\pm$ 0.021 & \cellcolor{highlight}2.14 $\pm$ 0.005 & \cellcolor{highlight}4.94 $\pm$ 0.003 & \cellcolor{highlight}6.11 $\pm$ 0.002 \\
HXE $\alpha$=0.6 & 34.68 $\pm$ 0.003 & 6.69 $\pm$ 0.007 & 2.32 $\pm$ 0.001 & 4.99 $\pm$ 0.005 & 6.13 $\pm$ 0.003 \\
\rowcolor{backgroundcolor}Soft-labels $\beta=30$ & 30.69 $\pm$ 0.125 & 6.99 $\pm$ 0.007 & 2.15 $\pm$ 0.008 & 4.95 $\pm$ 0.001 & \cellcolor{highlight}6.11 $\pm$ 0.001 \\
Soft-labels $\beta=4$ & 82.72 $\pm$ 0.079 & 7.54 $\pm$ 0.001 & 6.24 $\pm$ 0.005 & 6.94 $\pm$ 0.005 & 7.25 $\pm$ 0.002 \\
Chang et al. \cite{flamingo_2021_Chang} & 33.73 $\pm$ 0.033 & 6.93 $\pm$ 0.015 & 5.02 $\pm$ 0.007 & 2.34 $\pm$ 0.002 & 6.15 $\pm$ 0.001 \\
\rowcolor{backgroundcolor}\ours & \cellcolor{highlight}30.63 $\pm$ 0.007 & 6.97 $\pm$ 0.024 & \cellcolor{highlight}2.14 $\pm$ 0.008 & 4.95 $\pm$ 0.004 & \cellcolor{highlight}6.11 $\pm$ 0.001 \\
\addlinespace[0.1cm]
\hline
\end{tabular}
}
\end{center}
\caption{Results comparing top-$1$ error(\%) and hierarchical metrics on the test set of \textit{tieredImageNet-H}. The \textit{Top} block reports results without using CRM \cite{karthik_2021_ICLR} and the \textit{Bottom} block are reported using CRM. Rows highlighted with \colorbox{backgroundcolor}{lightgreen} are competitive methods in top-$1$ error (\%). Of these methods, we highlight the best performing entries for each metric with \colorbox{highlight}{green}.}
\label{tab:tiereimagenet}
\end{table*}

It is worth pointing out that Chang et al.'s \cite{flamingo_2021_Chang} method does not scale well with increasing number of hierarchical levels. For CIFAR-100 with six levels, the accuracy is competitive with cross-entropy, however, with both iNat and tieredImageNet-H, which have 8 and 13 levels, the top-$1$ error worsens. This is not unexpected as the feature vector is divided based on number of levels. While increasing the feature space may be a reasonable solution to maintain performance, it may not be straightforward to decide the feature vector size for each level, especially for hierarchies that may be skewed.
On the contrary, \ours is independent of the number of hierarchical levels used despite using hierarchical classifiers at each level. We also note that the CRM approach fails to improve Soft-labels $\beta$=4. This is perhaps because the label distribution is very flat for smaller $\beta$ values, leading to predictions with low confidence, which CRM could not help rectify.

\subsection{Coarse classification Accuracy}
\begin{figure*}[ht!]
   \subfloat[CIFAR-100 \label{fig:cifar-coarse}]{%
      \includegraphics[ width=0.32\textwidth]{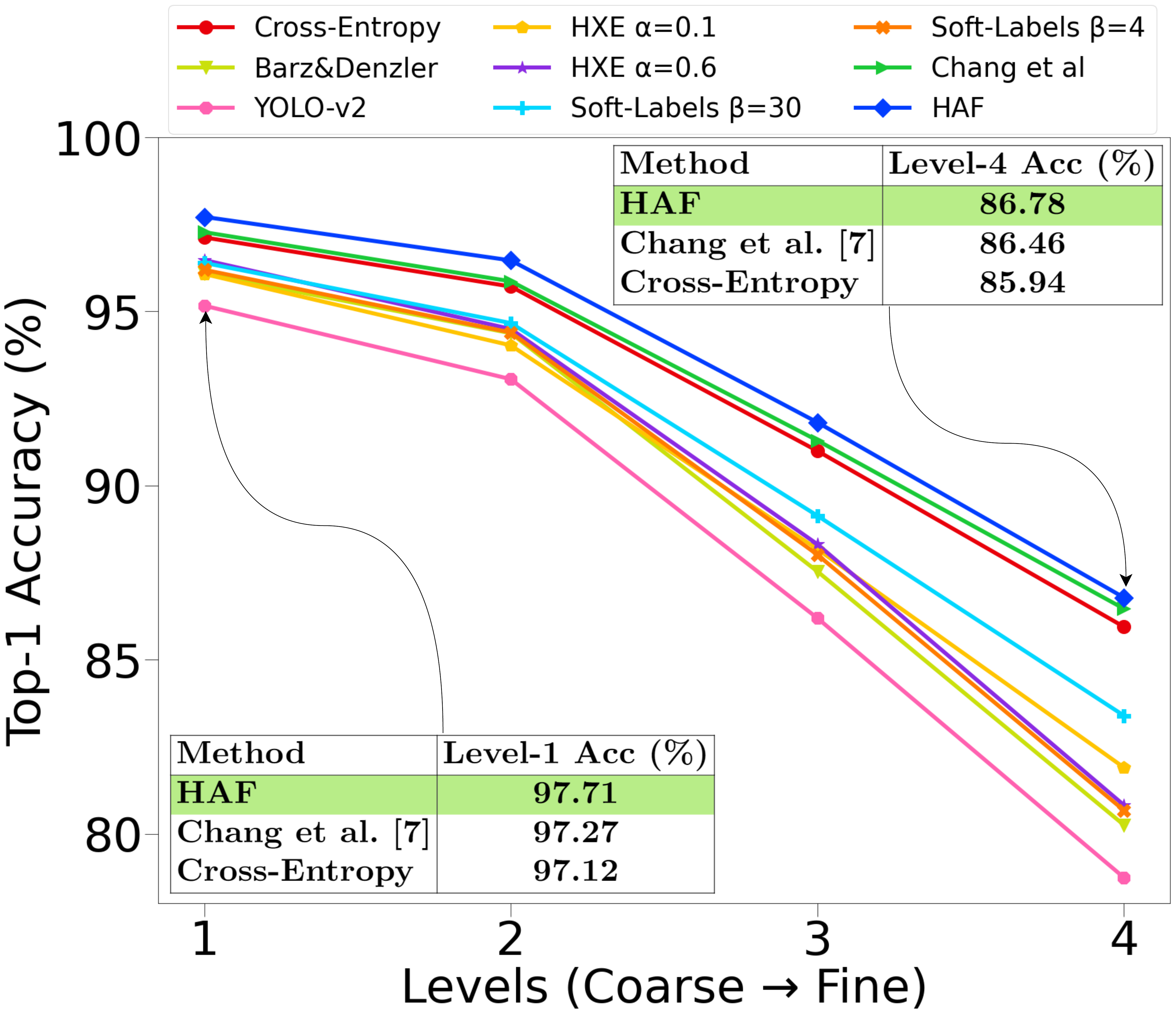}}
\hspace{\fill}
   \subfloat[iNaturalist19 \label{fig:inat-coarse} ]{%
      \includegraphics[ width=0.32\textwidth]{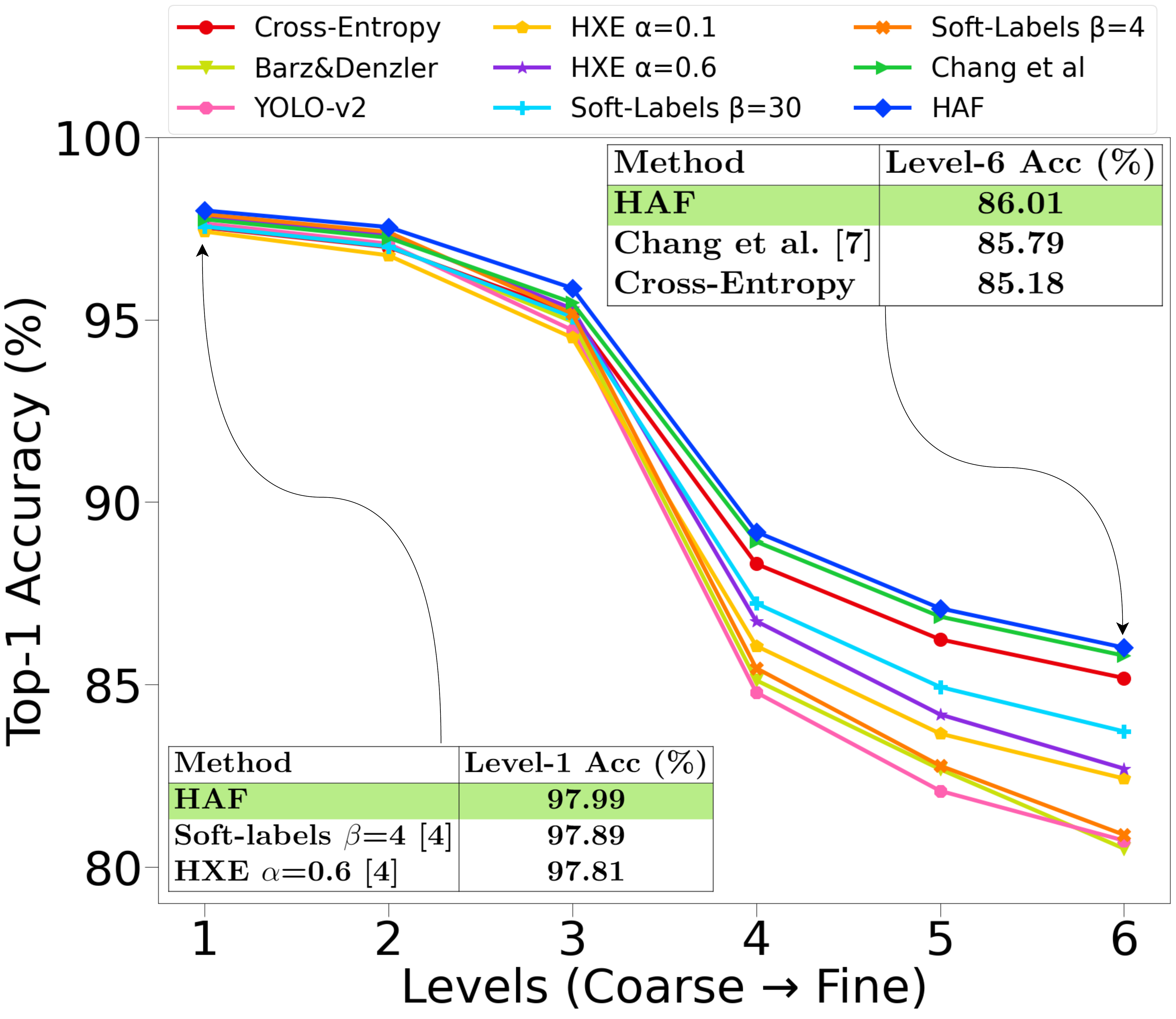}}
\hspace{\fill}
   \subfloat[tieredImageNet-H \label{fig:tiered-coarse}]{%
      \includegraphics[ width=0.32\textwidth]{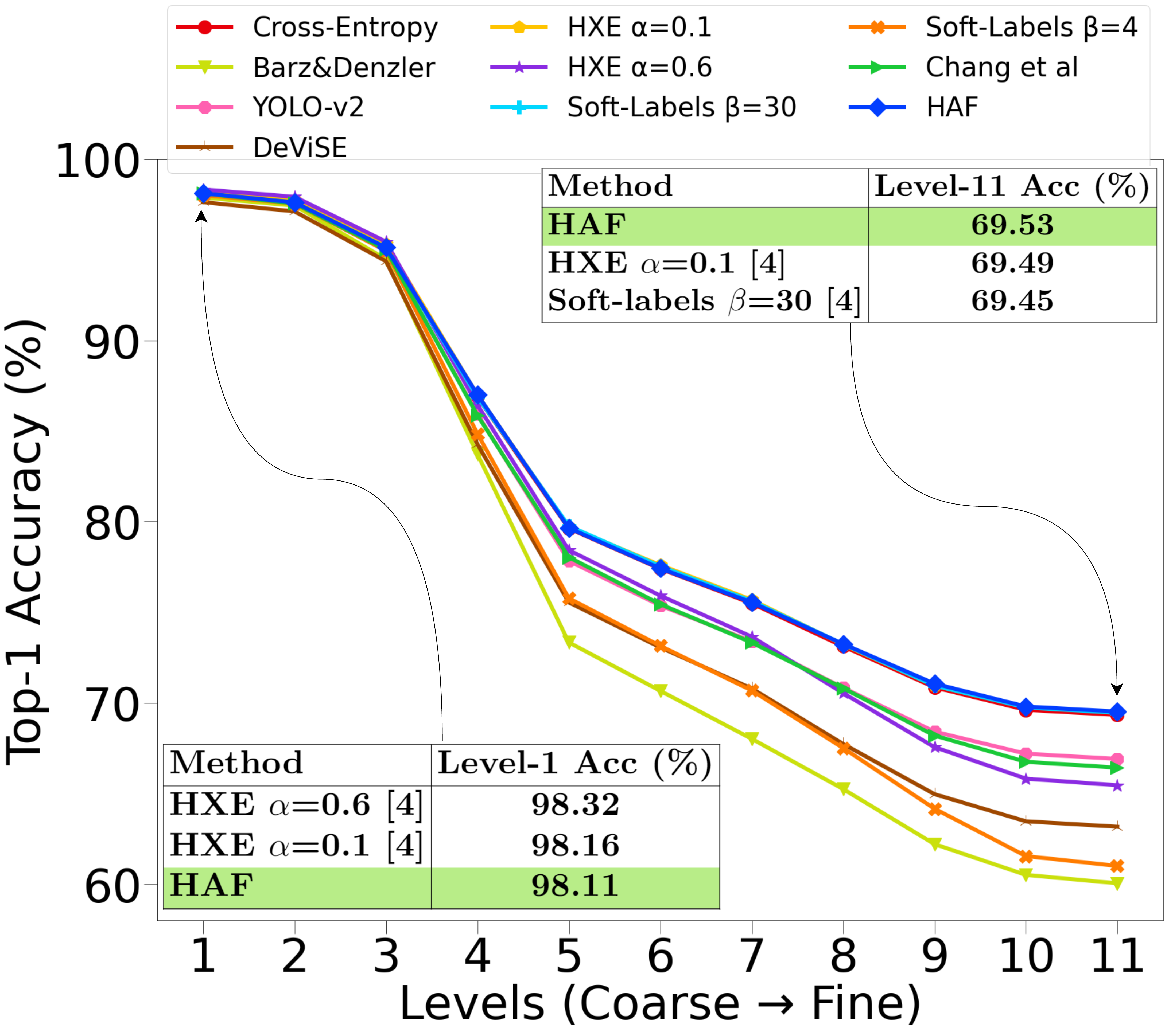}}\\
\caption{Coarse-level top-$1$ accuracy for each dataset. Level=$1$ is the coarsest level.}
    \label{coarse_level_wise_acc}
\end{figure*}

We also report comparisons over the coarse classification accuracy at all hierarchical levels. The learned feature representations guided with label hierarchies is expected to follow the structure of label hierarchies in the feature space. Such a feature space must restrict the confusions within their respective coarse classes, thereby, increasing the coarse-classification accuracy. 
We map the target labels and the predicted labels from the finest-level classifier to their respected coarse classes to evaluate the performance of the models on other hierarchical levels using coarse-classification accuracy. The results are reported in the Figure \ref{coarse_level_wise_acc}. On both CIFAR-100 and iNaturalist-19, \ours outperforms all the other baseline methods. On tieredImageNet-H, \ours has comparable performance with the Soft-labels $\beta$=30, HXE $\alpha$=0.1, and HXE $\alpha$=0.6.

\section{Analysis}\label{sec:analysis}

\subsection{Ablation Study}
In order to assess the contributions of each loss function used in our proposed approach, we present in Table \ref{tab:cifar100_inat-ablation-ours}, the results obtained with different variants of \ours on  CIFAR-100 and iNaturalist19 datasets respectively. It is evident that different variants of \ours perform slightly better than the cross-entropy baseline but \ours outperforms all the other variants. We can thus conclude that all the components of the loss function are significant and complementary for the overall performance of \ours. 


\begin{table*}[ht]
\begin{center}
\resizebox{0.9\textwidth}{!}{
\begin{tabular}{ c | cccc | ccccc } 
\hline
\addlinespace[0.2cm]
\multicolumn{1}{c}{\multirow{2}{*}{Method}} & \multicolumn{4}{c}{Loss function}                                                                                       & \multirow{2}{*}{Top-$1$ error($\downarrow$)} & \multirow{2}{*}{Mistakes severity($\downarrow$)} & \multirow{2}{*}{Hier Dist@1($\downarrow$)} & \multirow{2}{*}{Hier Dist@5($\downarrow$)} & \multirow{2}{*}{Hier Dist@20($\downarrow$)} \\
\multicolumn{1}{c}{}                        & \multicolumn{1}{c}{$L_{CE_{fine}}$} & \multicolumn{1}{c}{$L_{shc}$} & \multicolumn{1}{c}{$L_{gc}$} & \multicolumn{1}{c}{$L_m$} &                              \\
\addlinespace[0.1cm]
\hline
\addlinespace[0.1cm]
Cross-entropy & \checkmark & - & - & - & \cellcolor{highlight}22.11 & 2.37 & 0.52 & 2.24 & 3.16 \\
Variant of \ours & \checkmark & \checkmark & - & - & 22.70 & 2.36 & 0.54 & 1.61 & 2.78 \\
Variant of \ours & \checkmark & \checkmark & \checkmark & - & 22.35 & 2.32 & 0.52 & 1.66 & 2.87 \\
Variant of \ours & \checkmark & \checkmark & - & \checkmark & 22.12 & 2.24 & 0.5 & 1.44 & \cellcolor{highlight}2.61 \\
\ours & \checkmark & \checkmark & \checkmark & \checkmark & 22.25 & \cellcolor{highlight}2.22 & \cellcolor{highlight}0.49 & \cellcolor{highlight}1.40 & 2.64 \\
\addlinespace[0.1cm]
\hline
\hline
Cross-entropy & \checkmark & - & - & - & 36.48 & 2.39 & 0.87 & 1.97 & 3.25 \\
Variant of \ours & \checkmark & \checkmark & - & - & \cellcolor{highlight}36.23 & 2.34 & 0.85 & 1.73 & 2.81 \\
Variant of \ours & \checkmark & \checkmark & \checkmark & - & 36.60 & 2.32 & 0.85 & 1.71 & 2.73 \\
Variant of \ours & \checkmark & \checkmark & - & \checkmark & 36.34 & 2.31 & 0.84 & 1.76 & 2.91 \\
\ours & \checkmark & \checkmark & \checkmark & \checkmark &  36.47 & \cellcolor{highlight}2.27 & \cellcolor{highlight}0.83 & \cellcolor{highlight}1.62 & \cellcolor{highlight}2.56 \\
\addlinespace[0.1cm]
\hline
\end{tabular}
}
\end{center}
\caption{Ablative study comparing top-$1$ error(\%) and hierarchical metrics on the test sets of \textit{CIFAR-100} (top) and \textit{iNaturalist-19} (bottom).}
\label{tab:cifar100_inat-ablation-ours}
\end{table*}


\subsection{Mistakes Severity Plots}
\begin{figure*}[ht!]
   \subfloat[CIFAR-100 \label{fig:cifar-mistakes-severity}]{%
      \includegraphics[ width=0.32\textwidth]{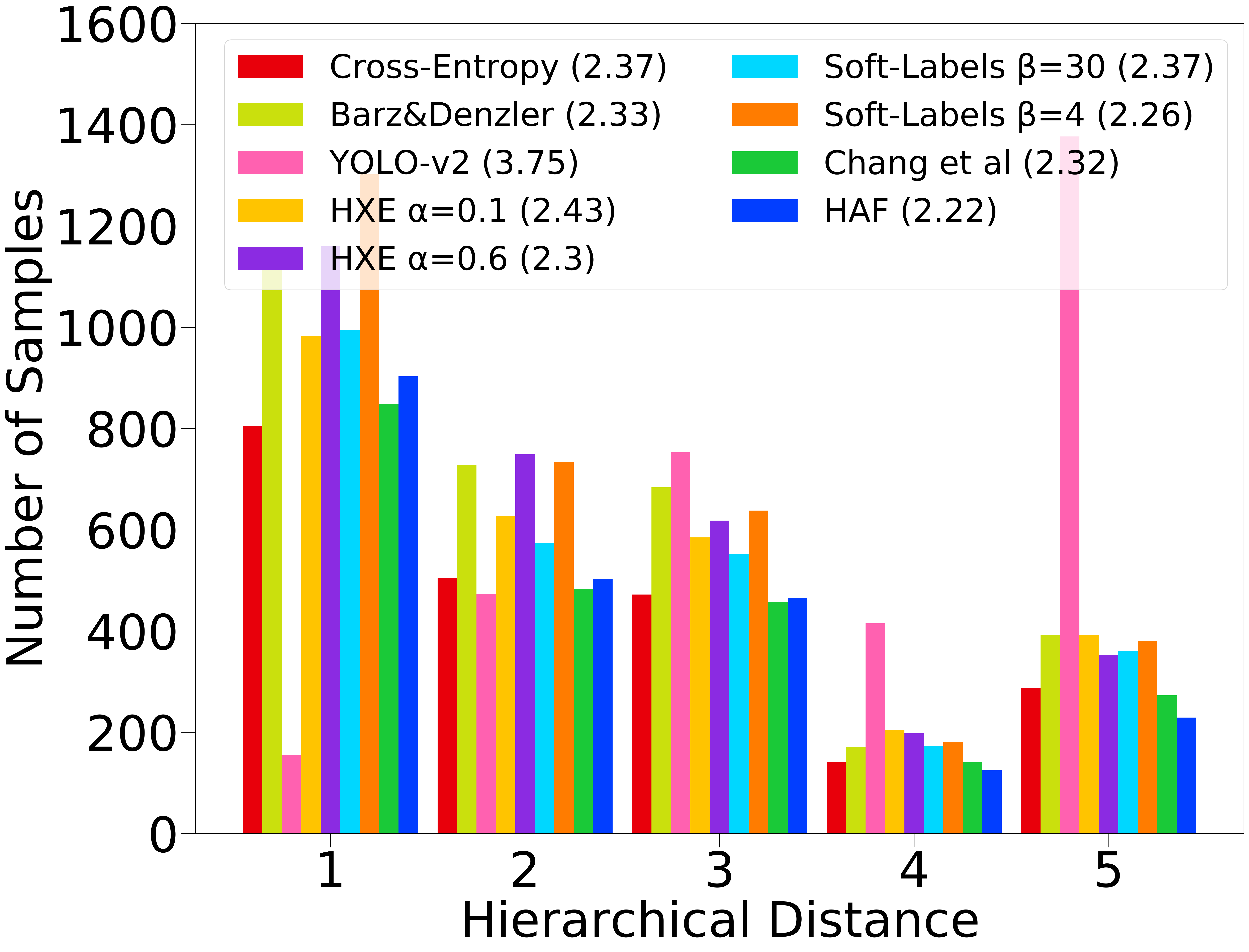}}
\hspace{\fill}
   \subfloat[iNaturalist19 \label{fig:inat-mistakes-severity} ]{%
      \includegraphics[ width=0.32\textwidth]{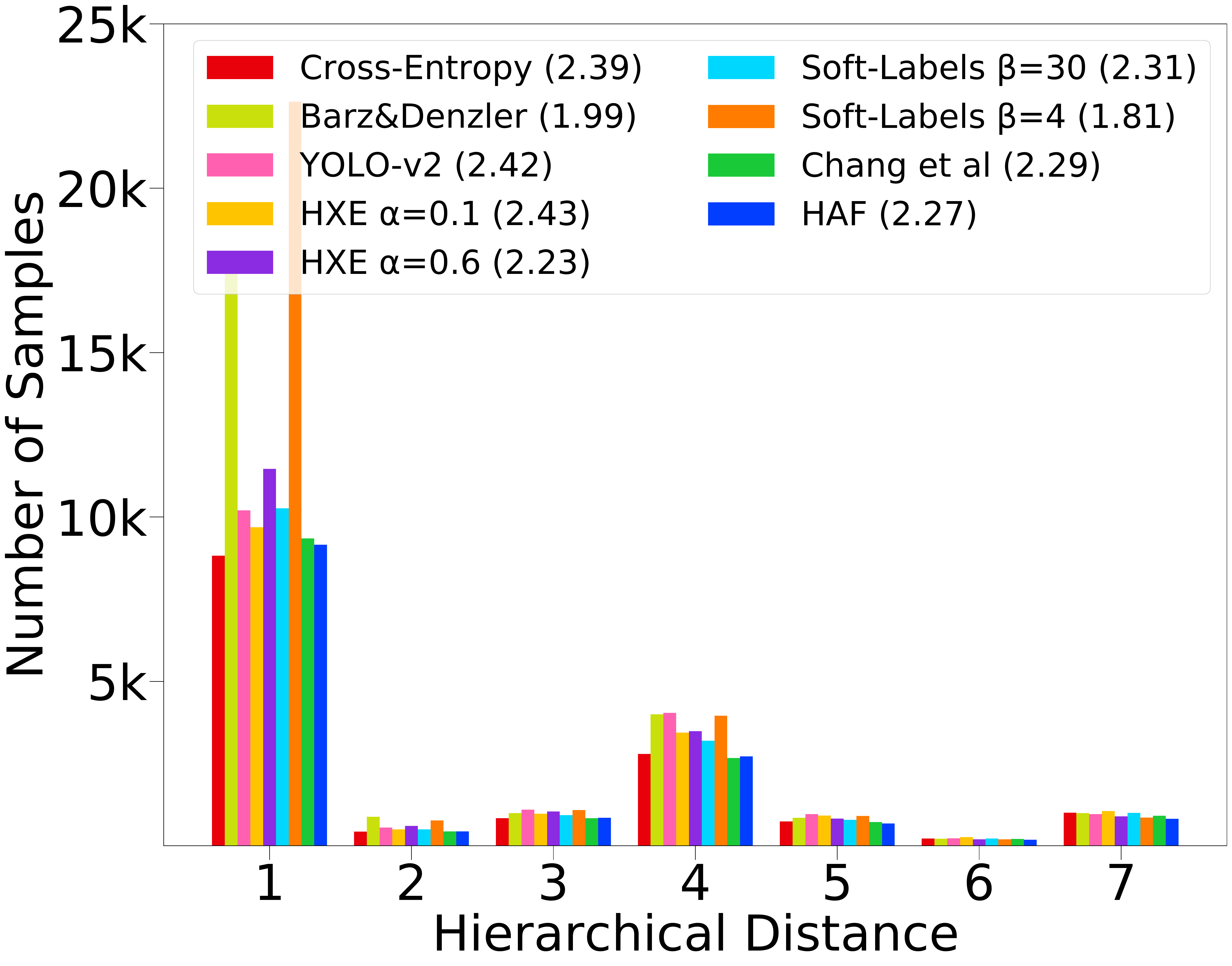}}
\hspace{\fill}
   \subfloat[tieredImageNet-H \label{fig:tiered-mistakes-severity}]{%
      \includegraphics[ width=0.32\textwidth]{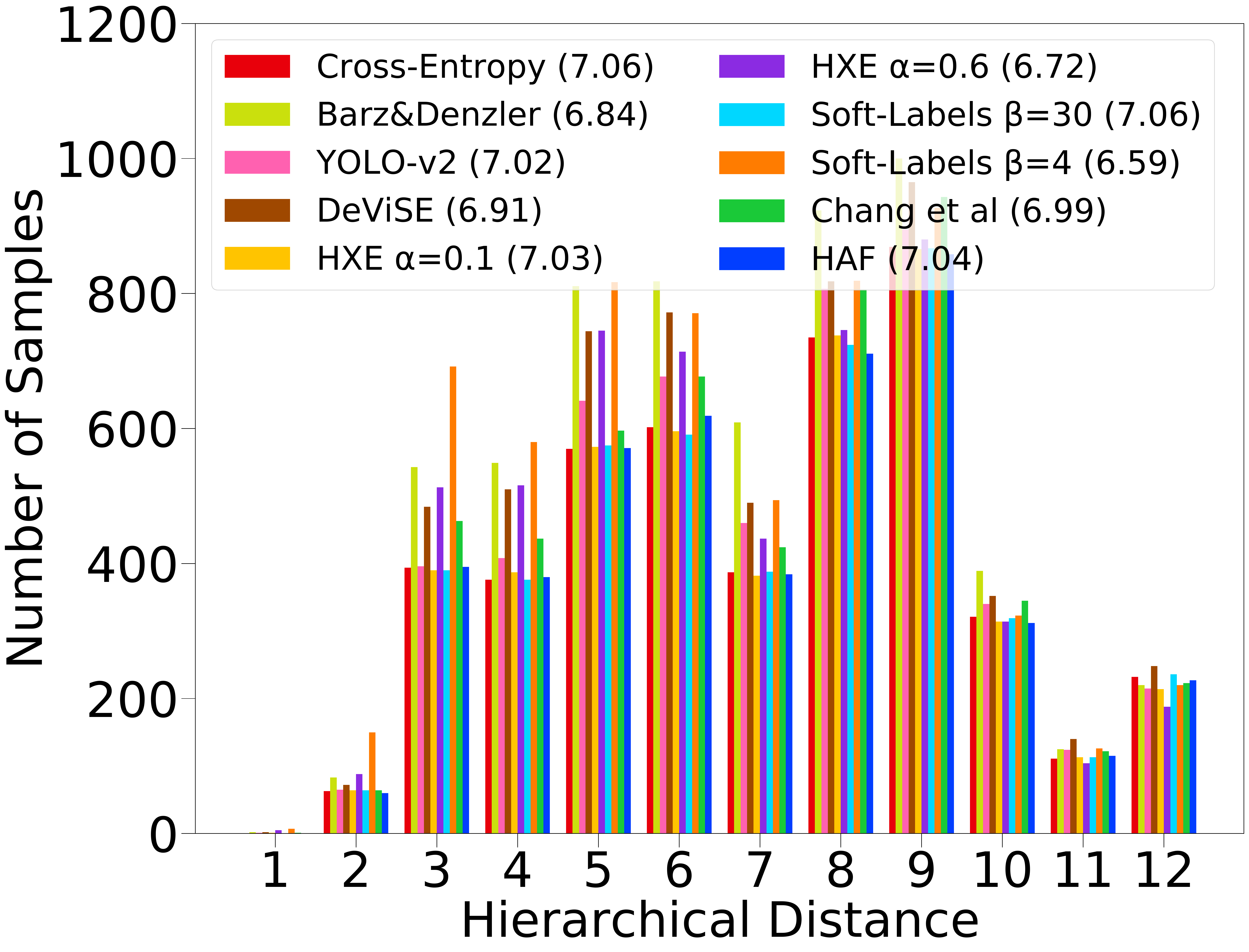}}\\
\caption{Mistakes severity plot showing distributions of mistakes at each level for each dataset. Numbers in the bracket denote the mistake severity of the method.}
    \label{fig:mistakes_severity}
\end{figure*}

We plot histograms to compare \ours with the baselines depicting the distribution of mistakes at different hierarchical levels. We present them for each dataset in Fig. \ref{fig:mistakes_severity}.
Mistakes at hierarchical distance $1$ refers to the mistakes with LCA=$1$. On CIFAR-100, \ours has the lowest mistake severity compared to all the methods and has number of mistakes comparable to cross-entropy at all levels except for level-$1$, where Chang et al. \cite{flamingo_2021_Chang} generates fewer mistakes. However, \ours has lesser number of high severity mistakes compared to Chang et al.\cite{flamingo_2021_Chang} which is a more desirable solution. On iNaturalist-19 dataset, soft-labels $\beta=4$, Barz \& Denzler, and HXE $\alpha=0.6$ has lower mistake severity as compared to \ours, but \ours makes lesser or nearly equal number of mistakes compared to these methods at all the hierarchical levels. On tieredImageNet-H, Barz \& Denzler, DeViSE, HXE $\alpha=0.6$, soft-labels $\beta=4$ has lower mistakes severity than \ours but much larger number of mistakes at every level. The metric `mistakes severity' alone does not give a complete picture of a method's ability to improve the mistakes. 

\subsection{Discussion: Hierarchical Metrics}

\begin{wrapfigure}[15]{R}{0.4\textwidth}
\centering{
    \includegraphics[width=0.39\textwidth]{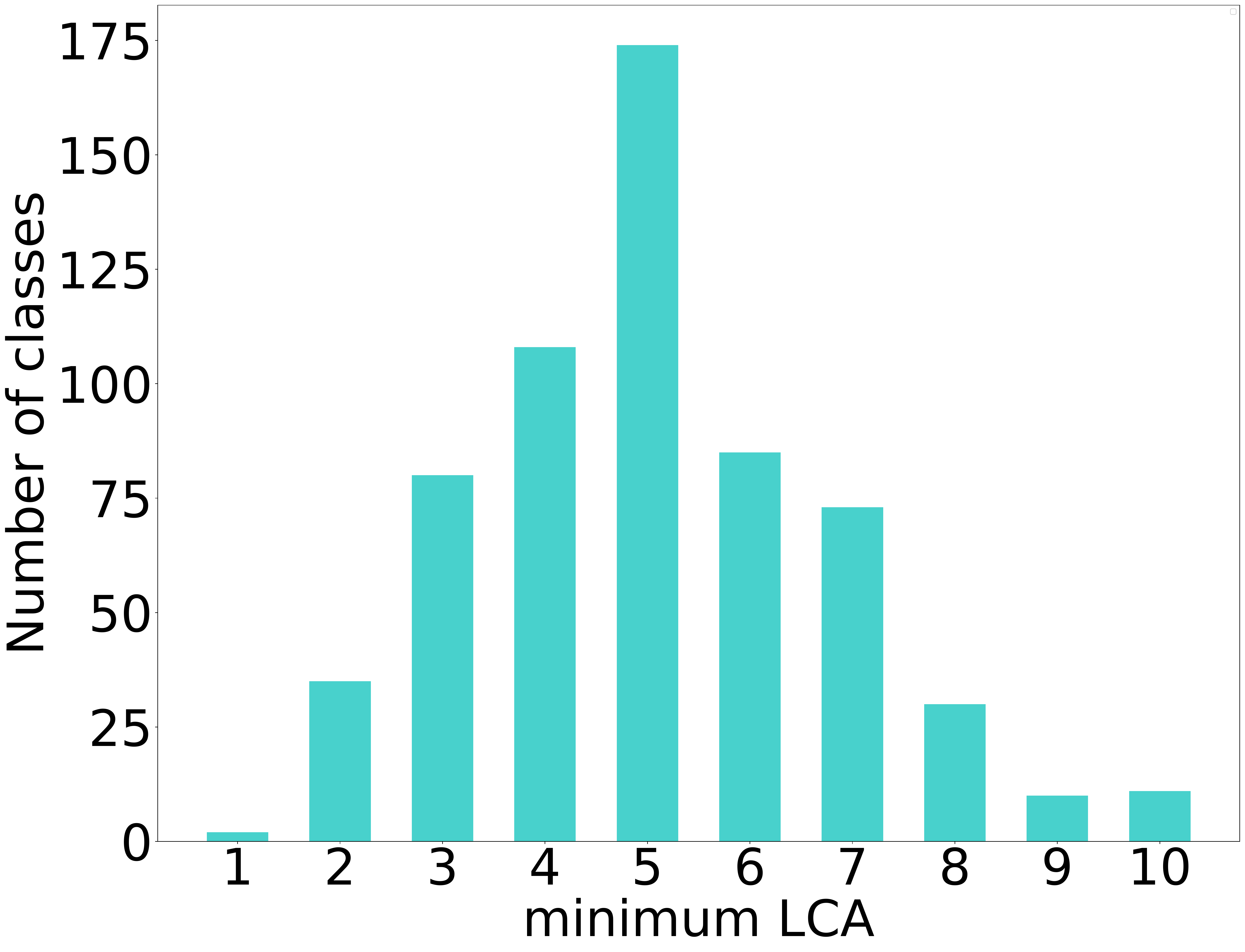}
     \caption{Number of classes with minimum LCA for tieredImageNet-H dataset.}
  \label{fig:tiered_plot_stats}
  }
\end{wrapfigure}

We discuss the inadequacy of the hierarchical metrics that have been proposed thus far. Figure \ref{fig:tiered_plot_stats} plots the histogram of the smallest possible LCA for all classes of the tieredImageNet-H dataset. Most classes have a minimum LCA greater than one, which indicates a skewed hierarchy tree, in turn explaining the high values of the hierarchical metrics in Table \ref{tab:tiereimagenet} across all methods. When these metrics are averaged over samples, the resulting change turns out to be very small, as observed by the reported standard deviations in Tables \ref{tab:cifar-100} - \ref{tab:tiereimagenet}. This problem of large values persists in all LCA-based metrics, and is dependent on the label hierarchy tree. 

As is depicted above in Figure \ref{fig:mistakes_severity}, mistakes severity favours the model with the reduction of \emph{average} LCA over the mistakes, implying that this metric may prefer a model with a large number of low-severity mistakes. Karthik et. al \cite{karthik_2021_ICLR} highlights the problems with mistakes severity. They overcome this drawback by using average hierarchical distance@$1$. 
However, we also note the problem with average hierarchical distance@$1$ metric. It is an average LCA distance of \textit{all} the samples from ground-truth to the top-1 predictions. This average includes as many zeros as the number of correct predictions (since the LCA distance for a correct prediction is 0). Therefore, it favours models that make fewer overall mistakes and thus fails to adequately capture the notion of a mistake's severity. 

An ideal method is the one that improves the mistakes severity metric while maintaining (or improving) the top-$1$ error, i.e., the sum of LCA of mistakes should reduce while maintaining (or improving) the total number of errors. On the CIFAR-100 dataset, we note that with a minimal drop in the top-$1$ accuracy, there is nearly $5\%$ improvement in reducing the sum of LCA of mistakes using \ours + CRM as compared to $2.17\%$ using cross-entropy + CRM. Similarly, on iNaturalist-19, \ours + CRM minimizes the sum of LCA of mistakes by 5.68\% compared to $2\%$ on cross-entropy + CRM. We present a more detailed analysis of these metrics in the supplementary material and defer the design for a more appropriate metric to measure mistake severity for future work.

\vspace{-5mm}
\section{Conclusion}\label{sec:conclusion}
In this paper, we introduced a novel approach to learn a hierarchy-aware feature space, which can preserve or improve the top-$1$ error and yet reduce the severity of mistakes. Our approach uses auxiliary classifiers at each level of the hierarchy that are trained by minimizing a Jensen-Shannon Divergence with target soft labels derived from finer-grained predictions of the samples. This training strategy regularizes the fine-grained classifier to make more consistent predictions with the coarser level classifiers, leading to a reduction in severity of mistakes. We further impose geometric consistency constraints between coarse and fine classifiers that leads to better alignment of the feature space distributions of the sub-classes with that of their super-classes. Without any additional hyperparameters, we simply trained our models with these loss functions and showed a reduction in mistake severity without trading off the top-$1$ error. We reported results from extensive experiments over three large datasets with varying levels of hierarchy and showed the strengths of our proposed method. We also presented an analysis of the commonly used hierarchical metrics and highlighted their limitations. We note that there exist recent works that leverage non-Euclidean spaces to learn appropriate embeddings for hierarchical data. However, much of the recent work on evaluating mistake severity is restricted to Euclidean feature spaces, and we present our analysis in the same space. Nonetheless, we conjecture that the nature of our contributions in this paper, i.e., losses that impose probabilistic and geometric constraints, would also extend to non-Euclidean spaces like hyperbolic feature spaces and would serve as a promising direction for future work. 



\section*{Acknowledgement}
Ashima Garg was supported by SERB, Govt. of India, under grant no. CRG/2020/ 006049. Depanshu Sani was supported by Google's AI for Social Good ``Impact Scholars'' program, 2021. Saket Anand gratefully acknowledges for the partial support from the Infosys Center for Artificial Intelligence at IIIT-Delhi. 

\newpage
\section*{Supplementary Material}
\section{Hierarchical cross-entropy (HXE) \cite{bertinetto_2020_CVPR}}
Here, we provide a derivation asserting our claims in the main text that the hierarchical cross-entropy (HXE) loss is a weighted combination of the cross-entropy loss applied at different hierarchical levels. 

Following the notations as given in \cite{bertinetto_2020_CVPR}. Define $p(C)$ is the categorical distribution over classes. The path from a leaf node $C$ to the root $R$ is $C^{(0)} = C, \ldots,  C^{(h)} = R$, the probability of class C can be factorised as 
\begin{gather}
    p(C) = \prod_{l=0}^{h-1} p(C^{(l)}|C^{(l+1)})
\end{gather}
Level $h$=0 is the finest-level in \cite{bertinetto_2020_CVPR}. Therefore, $p({C^{(h)}})=1$ at root level, and hence, last term is omitted in the above expression. Further, we use $L_{\text{CE}}^{h}$ to denote the cross-entropy at level-$h$. $y_{true}$ is equal to 1 if the true class is same as that of $C^k$. 

\begin{gather}
    L_{\text{CE}}^{0}(p, C) = -y_{\text{true}}*\log {p(C)}\\
    L_{\text{CE}}^{0}(p, C) = -y_{\text{true}}*\log {\big(p(C^0 | C^1).p(C^1 | C^2).p(C^2 | C^3) \ldots p(C^{(h-1)} | C^{(h)})\big)}
\end{gather}
The relation between the conditional probability and the cross-entropy is only valid when the probabilities of the true class are considered i.e. when $y_{true} = 1$. 
\begin{gather}
    L_{\text{CE}}^{0}(p, C) = -\big[\log p(C^0 | C^1) + \log p(C^1 | C^2) + \ldots + \log p(C^{(h-1)} | C^{(h)})\big]
\end{gather}
Similarly, 
\begin{gather}
    L_{\text{CE}}^{1}(p, C) = -\big[\log p(C^1 | C^2) + \log p(C^2 | C^3) \ldots + \log p(C^{(h-1)} | C^{(h)})\big]
\end{gather}
For any level $k$, the generalized equation can be written as: 
\begin{gather}
    L_{\text{CE}}^{(k)}(p, C) = -\big[\log p(C^{(k)} | C^{(k+1)}) + \log p(C^{(k+1)} | C^{(k+2)}) + \notag \\ \ldots + \log p(C^{(h-1)} | C^{(h)})\big]
\end{gather}

\begin{gather}
L_{\text{CE}}^{(k)}(p, C) = -\log p(C^{(k)} | C^{(k+1)}) + L_{\text{CE}}^{(k+1)}(p, C)\\
-\log p(C^{(k)} | C^{(k+1)}) = L_{\text{CE}}^{(k)}(p, C) - L_{\text{CE}}^{(k+1)}(p, C)\\
\log p(C^{(k)} | C^{(k+1)}) =  -\Big[ L_{\text{CE}}^{(k)}(p, C) - L_{\text{CE}}^{(k+1)}(p, C) \Big]
\label{eqn:pCk_pCk+1}
\end{gather}
Hierarchical cross-entropy (HXE) \cite{bertinetto_2020_CVPR} loss is given as: 
\begin{gather}
    L_{\text{HXE}}(p, C) = -\sum_{l=0}^{h-1} \lambda^{(C^l)} \log p (C^{(l)} | C^{(l+1)})
\end{gather}
\begin{gather}
    L_{\text{HXE}}(p, C) = -\big[ \lambda^{(C^0)} \log (p^{C^0}|p^{C^1}) + \lambda^{(C^1)} \log (p^{C^1}|p^{C^2}) + \ldots \notag\\ + \lambda^{(C^{(h-1)})}\log (p^{(C^{(h-1)})}|p^{(C^{(h)})}) \big]
\end{gather}
Substituting the value of $\log p(C^{(k}) | C^{(k+1)})$ from Eq. \ref{eqn:pCk_pCk+1}

\begin{gather}
 L_{\text{HXE}}(p, C) = \lambda^{(C^0)} \left[ L_{\text{CE}}^{0}(p, C) - L_{\text{CE}}^{1}(p, C) \right] +\lambda^{(C^1)} \left[ L_{\text{CE}}^{1}(p, C) - L_{\text{CE}}^{2}(p, C) \right] \notag\\ \qquad\qquad\qquad\qquad + \ldots +\lambda^{(C^{(h-1)})} \left[ L_{\text{CE}}^{(h-1)}(p, C) - L_{\text{CE}}^{(h)}(p, C) \right]
\end{gather}

\begin{gather}
 L_{\text{HXE}}(p, C) = \lambda^{(C^0)} L_{\text{CE}}^{0}(p, C) + \left[ \lambda^{(C^1)} - \lambda^{(C^0)}\right] L_{\text{CE}}^{1}(p, C) + \notag\\ \qquad\qquad\qquad\qquad \left[ \lambda^{(C^2)} - \lambda^{(C^1)}\right] L_{\text{CE}}^{2}(p, C) + \ldots - \lambda^{(C^{(h-1)})} L_{\text{CE}}^{(h)}(p, C)
\end{gather}

Thus, we have obtained the desired expression where $L_{\text{HXE}}$ as weighted sum of cross-entropy loss at different levels of the hierarchy. 
\section{Analysis of Hierarchical Metrics}

\begin{figure*}[h!]
   \subfloat[CIFAR-100 \label{fig:per_class_lca_cifar-100}]{%
      \includegraphics[ width=0.31\textwidth]{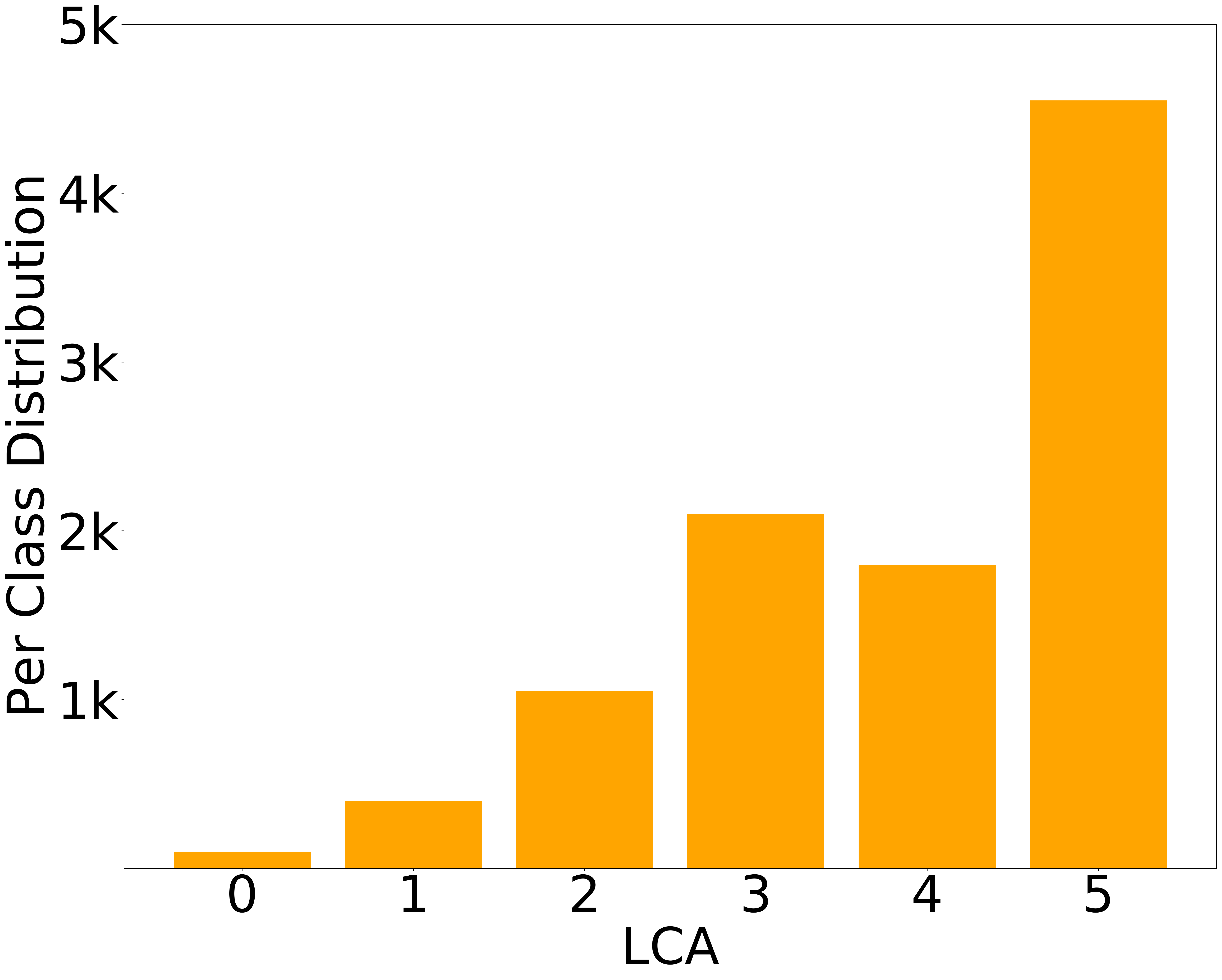}}
\hspace{\fill}
   \subfloat[iNaturalist19 \label{fig:per_class_lca_inat} ]{%
      \includegraphics[ width=0.31\textwidth]{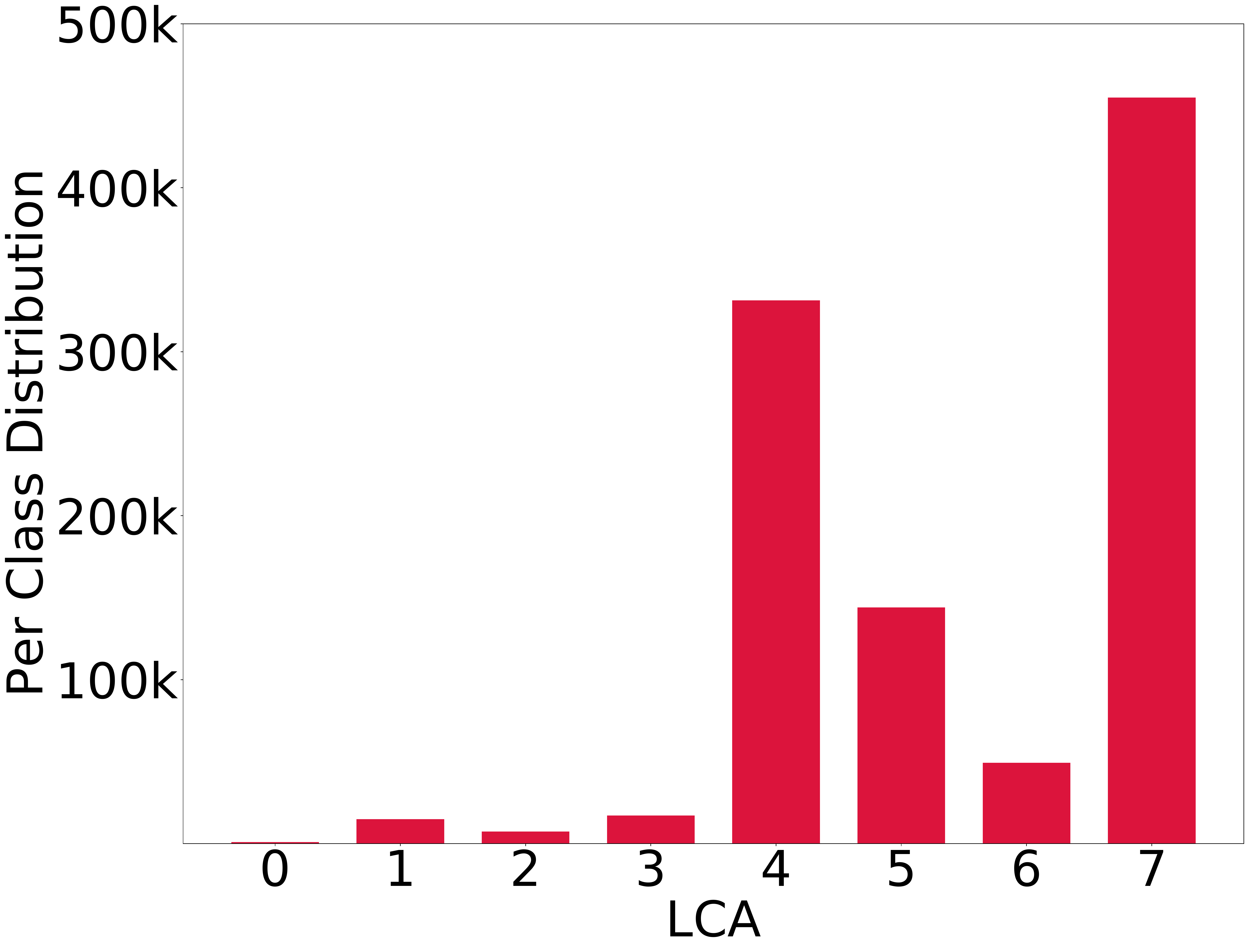}}
\hspace{\fill}
   \subfloat[tieredImageNet-H \label{fig:per_class_lca_tiered}]{%
      \includegraphics[ width=0.31\textwidth]{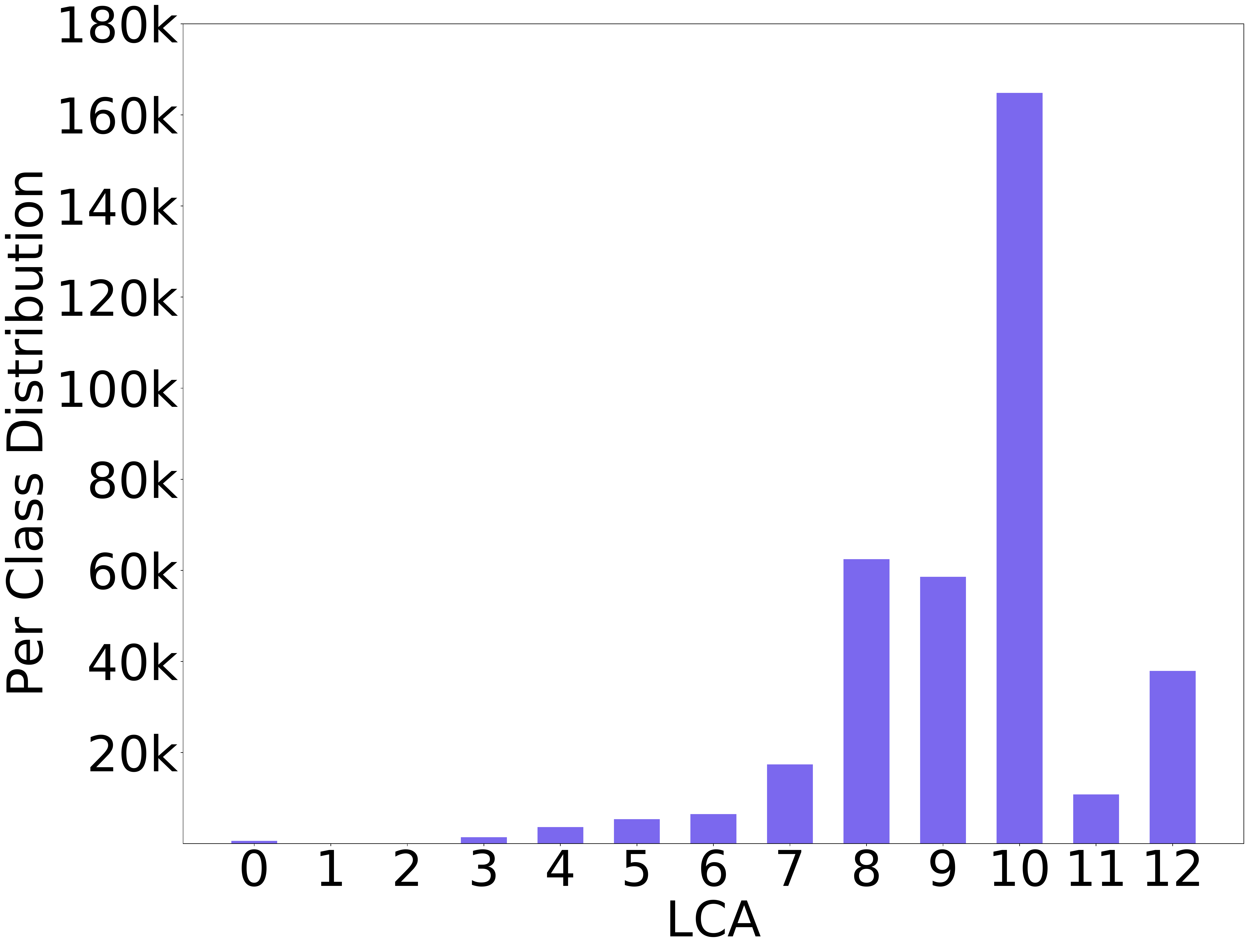}}\\
\caption{Per class distribution of LCA for each dataset.}
    \label{fig:per_class_distribution_LCA}
\end{figure*}

To plot Figure \ref{fig:per_class_distribution_LCA}, we first create a symmetric $|\calC|\times|\calC|$ LCA matrix where $|\calC|$ is the total number of fine-grained classes and each entry LCA($i$, $j$) denotes the LCA between class $i$ and class $j$. 
For every class, we compute the count of distinct LCA values using this matrix and sum them up for all the classes to plot this distribution for all the datasets. 
The misclassified samples will likely introduce the errors with LCA value at the peak of the plots. This plot shows the skewness of the hierarchy tree, resulting in larger values of the hierarchical metrics. 
\begin{table}
\begin{subtable}[c]{0.5\textwidth}
\centering
\resizebox{0.8\textwidth}{!}{
\begin{tabular}{l | c c}
Method & \begin{tabular}[c]{@{}c@{}}LCA sum\end{tabular} & \begin{tabular}[c]{@{}c@{}}Total mistakes\end{tabular} \\\hline
 \rowcolor{lightgray} Cross-Entropy & 5242.67 & 2227 \\
YOLO-v2 \cite{redmon_2017_yolo_CVPR} & 11917.33 (-127.31) & 3203.33 (-43.84) \\
HXE $\alpha$=0.1 \cite{bertinetto_2020_CVPR} & 6893.67 (-31.49) & 2840.67 (-27.56) \\
HXE $\alpha$=0.6 \cite{bertinetto_2020_CVPR} & 6965.33 (-32.86) & 3041.67 (-36.58) \\
soft-labels $\beta$=4 \cite{bertinetto_2020_CVPR} & 7100.33 (-35.43) & 3215.33 (-44.38) \\
soft-labels $\beta$=30 \cite{bertinetto_2020_CVPR} & 6411 (-22.29) & 2699.33 (-21.21) \\
Chang et al. \cite{flamingo_2021_Chang} & 5081.33 (3.08) & \cellcolor{highlight}\textbf{2194 (1.48)} \\
\ours & \cellcolor{highlight}\textbf{4992.67 (4.77)} & 2227 (0) \\
\hdashline & \\
Cross-Entropy + CRM \cite{karthik_2021_ICLR} & 5128.67 (2.17) & \cellcolor{highlight}\textbf{2223 (0.18)}\\
\ours + CRM & \cellcolor{highlight}\textbf{4970 (5.02)} & 2231.33 (-0.19)
\end{tabular}
}
\subcaption{CIFAR-100}
\end{subtable}
\begin{subtable}[c]{0.5\textwidth}
\centering
\resizebox{0.8\textwidth}{!}{
\begin{tabular}{l | c c}
Method & \begin{tabular}[c]{@{}c@{}}LCA sum \end{tabular} & \begin{tabular}[c]{@{}c@{}}Total mistakes\end{tabular} \\\hline
 \rowcolor{lightgray} Cross-Entropy & 35458.33 & 14846 \\
YOLO-v2 \cite{redmon_2017_yolo_CVPR} & 43814.33 (-23.57) & 18074.33 (-21.75)  \\
HXE $\alpha$=0.1 \cite{bertinetto_2020_CVPR} & 40884.67 (-15.3) & 16898 (-13.82) \\
HXE $\alpha$=0.6 \cite{bertinetto_2020_CVPR} & 41388 (-16.72) & 18516 (-24.72) \\
Soft-labels $\beta$=4 \cite{bertinetto_2020_CVPR} & 55241(-55.79) & 30432 (-104.98) \\
Soft-labels $\beta$=30 \cite{bertinetto_2020_CVPR} & 39443 (-11.24) & 16976 (-14.35) \\
Chang et al. \cite{flamingo_2021_Chang} & 34631.67 (2.33) & 15168 (-2.17) \\
\ours & \cellcolor{highlight}\textbf{33732 (4.54)} & \cellcolor{highlight}\textbf{14859 (0.13)} \\
\hdashline & \\
Cross-Entropy + CRM \cite{karthik_2021_ICLR} & 34724 (2.07) & 14872.33 (-0.18) \\
\ours + CRM & \cellcolor{highlight}\textbf{33446 (5.68)} & \cellcolor{highlight}\textbf{14859 (-0.09)}
\end{tabular}
}

\subcaption{iNaturalist-19}
\end{subtable}
\caption{LCA sum i.e. sum of LCA of mistakes and total mistakes on CIFAR-100 and iNaturalist-19. The values reported are the average of three different seeds. }
\label{tab:sum_LCA_mistakes}
\end{table}

An ideal method is the one that improves the mistakes severity metric while maintaining (or improving) the top-$1$ accuracy, i.e., the \textit{LCA sum} which is the sum of LCA of misclassified samples, should reduce while maintaining (or improving) the total number of errors. We analyze the \textit{LCA sum} parallel to the total number of mistakes for each of the baseline methods on CIFAR-100 and iNaturalist-19 dataset in the Table \ref{tab:sum_LCA_mistakes}. The numbers in the parentheses of the column of the \textit{LCA sum} denote the percentage improvement in reducing the LCA sum compared to the baseline cross-entropy. While the numbers in the parentheses of the column of the \textit{Total mistakes} indicate the percentage improvement in reducing the total number of errors compared to the baseline cross-entropy. 

\section{Mistakes severity using CRM}

\begin{figure*}[ht!]
   \subfloat[CIFAR-100 \label{fig:cifar-mistakes-severity-test}]{%
      \includegraphics[ width=0.32\textwidth]{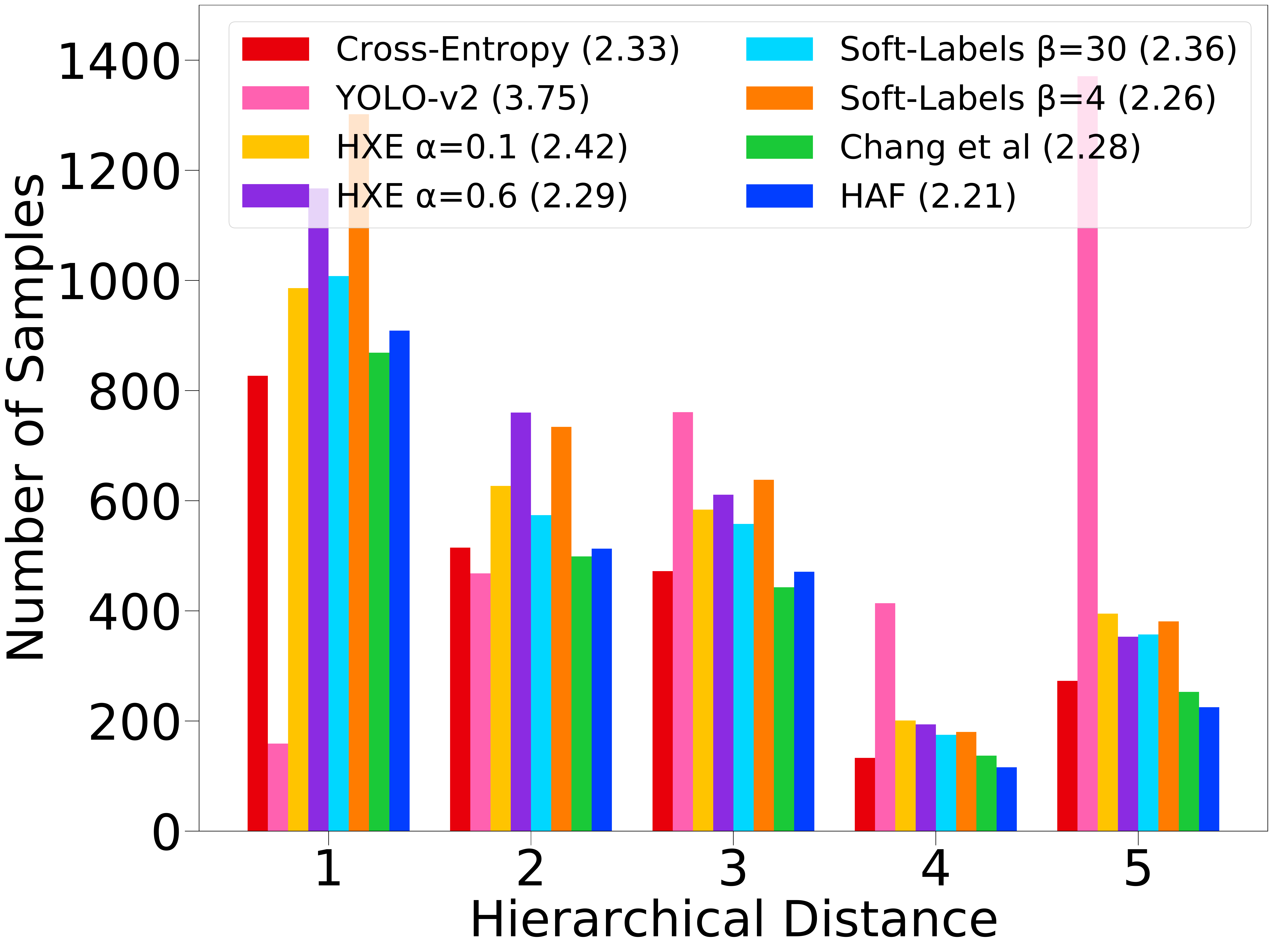}}
\hspace{\fill}
   \subfloat[iNaturalist19 \label{fig:inat-mistakes-severity-test} ]{%
      \includegraphics[ width=0.32\textwidth]{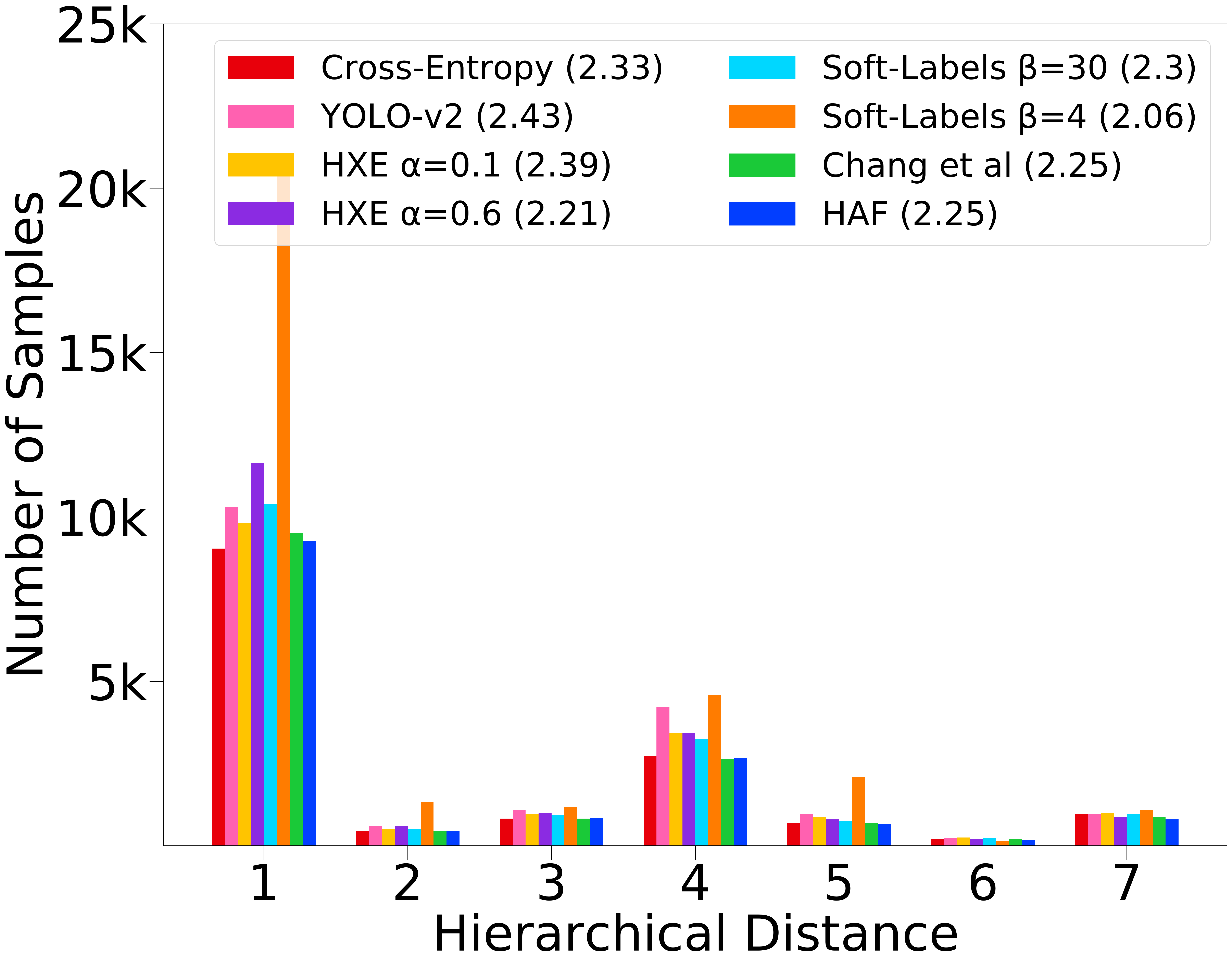}}
\hspace{\fill}
   \subfloat[tieredImageNet-H \label{fig:tiered-mistakes-severity-test}]{%
      \includegraphics[ width=0.32\textwidth]{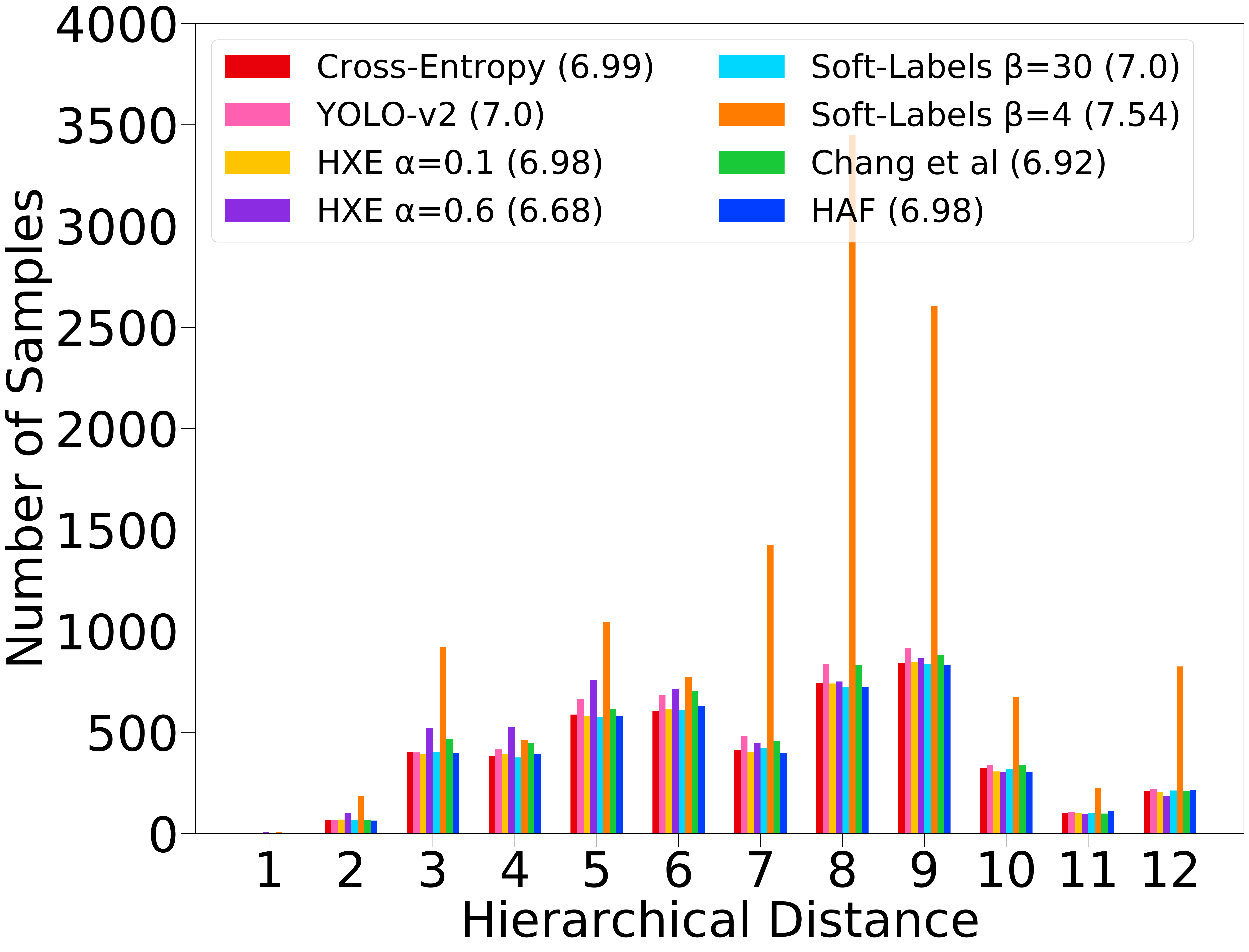}}\\
\caption{Mistakes severity plot showing distributions of mistakes at each level for each dataset when CRM \cite{karthik_2021_ICLR} is used for evaluation. Numbers in the bracket denote the mistake severity of the method.}
    \label{fig:mistakes_severity_CRM}
\end{figure*}

We plot the distribution of mistakes for methods when evaluated using CRM in Figure \ref{fig:mistakes_severity_CRM}. Our observations are consistent with Section 5.2 (main text) on all the datasets. CRM benefits most of the methods except Soft-labels $\beta=4$. The performance of Soft-labels $\beta=4$ drops when evaluated using CRM. The same reason stated earlier is that the label distribution is flat for smaller $\beta$ values, leading to predictions with low confidence.

\section{Coarse classification Accuracy}

In Table \ref{tab:cifar100_coarse}, \ref{tab:inat_coarse} and \ref{tab:tiered_coarse} we present the comparisons of \ours with the baselines on coarse-classification accuracy across all hierarchical levels on CIFAR-100, iNaturalist-19 and tieredImageNet-H respectively. L1 refers to Level-1 and is the coarsest-level. We report the coarse-classification accuracy with and without using CRM at test-time. To obtain coarse-classification accuracy, we map the target labels and the predicted labels obtained from the finest-level classifier to their respected coarse classes.
We highlight the best entries in each of the column with \colorbox{highlight}{green}. Clearly, CIFAR-100 and iNaturalist-19 surpasses all other methods on both evaluation methods with and without using CRM. On tieredImageNet-H, \ours outperforms other methods towards finer-levels. 

\begin{table*}[h!]
\small
\begin{center}
\resizebox{0.6\textwidth}{!}{
    \begin{tabular}{l cccc}
    \hline Method & L1 & L2 & L3 & L4\\\hline
    Cross-Entropy & ~~97.12~~ & ~~95.71~~ & ~~90.99~~ & ~~85.94~~\\
    Barz \& Denzler \cite{barz_2019_WACV} & ~~96.08~~ & ~~94.37~~ & ~~87.53~~ & ~~80.25~~ \\
    YOLO-v2 \cite{redmon_2017_yolo_CVPR} & ~~95.16~~ & ~~93.05~~ & ~~86.19~~ & ~~78.74~~\\
    HXE $\alpha$=0.1 \cite{bertinetto_2020_CVPR} & ~~96.07~~ & ~~94.02~~ & ~~88.17~~ & ~~81.90~~ \\
    HXE $\alpha$=0.6 \cite{bertinetto_2020_CVPR} & ~~96.47~~ & ~~94.49~~ & ~~88.31~~ & ~~80.82~~\\
    Soft-labels $\beta$=30 \cite{bertinetto_2020_CVPR} & ~~96.39~~ & ~~94.66~~ & ~~89.13~~ & ~~83.39~~\\
    Soft-labels $\beta$=4 \cite{bertinetto_2020_CVPR} & ~~96.19~~ & ~~94.39~~ & ~~88.01~~ & ~~80.67~~\\
    Chang et al. \cite{flamingo_2021_Chang} & ~~97.27~~ & ~~95.86~~ & ~~91.29~~ & ~~86.46~~\\
    \ours & ~~\cellcolor{highlight}97.71~~ & ~~\cellcolor{highlight}96.46~~ & ~~\cellcolor{highlight}91.81~~ & ~~\cellcolor{highlight}86.78~~ \\
    \hline
    Cross-Entropy \cite{karthik_2021_ICLR} & ~~97.27~~ & ~~95.94~~ & ~~91.22~~ & ~~86.07~~\\
    YOLO-v2 \cite{redmon_2017_yolo_CVPR} & ~~95.24~~ & ~~93.24~~ & ~~86.39~~ & ~~78.86~~\\
    HXE $\alpha$=0.1 \cite{bertinetto_2020_CVPR} & ~~96.05~~ & ~~94.04~~ & ~~88.20~~ & ~~81.93~~\\
    HXE $\alpha$=0.6 \cite{bertinetto_2020_CVPR} & ~~96.47~~ & ~~94.53~~ & ~~88.42~~ & ~~80.82~~\\
    Soft-labels $\beta$=30 \cite{bertinetto_2020_CVPR} & ~~96.43~~ & ~~94.68~~ & ~~89.10~~ & ~~83.36~~\\
    Soft-labels $\beta$=4 \cite{bertinetto_2020_CVPR} & ~~96.17~~ & ~~94.50~~ & ~~88.28~~ & ~~80.34~~\\
    Chang et al. \cite{flamingo_2021_Chang} & ~~97.47~~ & ~~96.10~~ & ~~91.67~~ & ~~86.68~~\\
    \ours & ~~\cellcolor{highlight}97.75~~ & ~~\cellcolor{highlight}96.59~~ & ~~\cellcolor{highlight}91.88~~ & ~~\cellcolor{highlight}86.75~~\\
    \hline
    \end{tabular}
}
\end{center}
\caption{Coarse-classification accuracy results on the test set of \textit{CIFAR-100}. The \textit{Top} block reports results without using CRM \cite{karthik_2021_ICLR} and the \textit{Bottom} block reports results using CRM on all coarse-levels from level-1(Coarse (L1)) to and level-4(Fine (L4)).}
\label{tab:cifar100_coarse}
\end{table*}

\begin{table}
\small
\begin{center}
\resizebox{0.7\textwidth}{!}{
    \begin{tabular}{l cccccc}
    \hline Method & L1 & L2 & L3 & L4 & L5 & L6\\\hline
    Cross-Entropy & ~~97.52~~ & ~~96.99~~ & ~~95.16~~ & ~~88.31~~ & ~~86.24~~ & ~~85.18~~ \\
    Barz \& Denzler \cite{barz_2019_WACV} & ~~97.56~~ & ~~97.03~~ & ~~94.94~~ & ~~85.12~~ & ~~82.68~~ & ~~80.50~~ \\
    YOLO-v2 \cite{redmon_2017_yolo_CVPR} & ~~97.64~~ & ~~97.08~~ & ~~94.71~~ & ~~84.79~~ & ~~82.09~~ & ~~80.72~~ \\
    HXE $\alpha$=0.1 \cite{bertinetto_2020_CVPR} & ~~97.41~~ & ~~96.76~~ & ~~94.51~~ & ~~86.06~~ & ~~83.65~~ & ~~82.42~~ \\
    HXE $\alpha$=0.6 \cite{bertinetto_2020_CVPR} & ~~97.81~~ & ~~97.32~~ & ~~95.30~~ & ~~86.74~~ & ~~84.18~~ & ~~62.89~~ \\
    Soft-labels $\beta$=30 \cite{bertinetto_2020_CVPR} & ~~97.55~~ & ~~97.01~~ & ~~95.07~~ & ~~87.22~~ & ~~84.93~~ & ~~83.71~~ \\
    Soft-labels $\beta$=4 \cite{bertinetto_2020_CVPR} & ~~97.89~~ & ~~97.40~~ & ~~95.17~~ & ~~85.45~~ & ~~82.77~~ & ~~80.88~~ \\
    Chang et al. \cite{flamingo_2021_Chang} & ~~97.75~~ & ~~97.24~~ & ~~95.47~~ & ~~88.91~~ & ~~86.86~~ & ~~85.79~~ \\
    \ours & ~~\cellcolor{highlight}97.99~~ & ~~\cellcolor{highlight}97.54~~ & ~~\cellcolor{highlight}95.86~~ & ~~\cellcolor{highlight}89.18~~ & ~~\cellcolor{highlight}87.09~~ & ~~\cellcolor{highlight}86.01~~ \\
    \hline
    Cross-Entropy \cite{karthik_2021_ICLR} & ~~97.62~~ & ~~97.14~~ & ~~95.43~~ & ~~88.73~~ & ~~86.70~~ & ~~85.60~~ \\
    YOLO-v2 \cite{redmon_2017_yolo_CVPR} & ~~97.63~~ & ~~97.06~~ & ~~94.70~~ & ~~84.33~~ & ~~81.62~~ & ~~80.17~~ \\
    HXE $\alpha$=0.1 \cite{bertinetto_2020_CVPR} & ~~97.55~~ & ~~96.92~~ & ~~94.79~~ & ~~86.37~~ & ~~83.98~~ & ~~82.72~~ \\
    HXE $\alpha$=0.6 \cite{bertinetto_2020_CVPR} & ~~97.83~~ & ~~97.34~~ & ~~95.37~~ & ~~86.97~~ & ~~84.50~~ & ~~83.01~~ \\
    Soft-labels $\beta$=30 \cite{bertinetto_2020_CVPR} & ~~97.61~~ & ~~97.06~~ & ~~95.19~~ & ~~87.24~~ & ~~84.95~~ & ~~83.72~~ \\
    Soft-labels $\beta$=4 \cite{bertinetto_2020_CVPR} & ~~97.31~~ & ~~96.93~~ & ~~91.80~~ & ~~80.51~~ & ~~77.60~~ & ~~74.30~~ \\
    Chang et al. \cite{flamingo_2021_Chang} & ~~97.86~~ & ~~97.37~~ & ~~95.68~~ & ~~89.22~~ & ~~87.19~~ & ~~86.11~~ \\
    \ours & ~~\cellcolor{highlight}98.02~~ & ~~\cellcolor{highlight}97.58~~ & ~~\cellcolor{highlight}95.96~~ & ~~\cellcolor{highlight}89.38~~ & ~~\cellcolor{highlight}87.30~~ & ~~\cellcolor{highlight}86.20~~ \\
    \hline
    \end{tabular}
}
\end{center}
\caption{Coarse-classification accuracy results on the test set of \textit{iNaturalist-19}. The \textit{Top} block reports results without using CRM \cite{karthik_2021_ICLR} and the \textit{Bottom} block reports results using CRM on all coarse-levels from level-1(Coarse (L1)) to and level-6(Fine (L6)).}
\label{tab:inat_coarse}
\end{table}

\begin{table}
\small
\begin{center}
\resizebox{1.0\textwidth}{!}{
    \begin{tabular}{l ccccccccccc}
    \hline Method & L1 & L2 & L3 & L4 & L5 & L6 & L7 & L8 & L9 & L10 & L11\\\hline
    Cross-Entropy & ~~98.07~~ & ~~97.62~~ & ~~95.17~~ & ~~86.88~~ & ~~79.62~~ & ~~77.40~~ & ~~75.49~~ & ~~73.12~~ & ~~70.86~~ & ~~69.62~~ & ~~69.34~~ \\
    Barz \& Denzler \cite{barz_2019_WACV} & 97.87 & 97.43 & 94.47 & 83.73 & 73.34 & 70.67 & 68.03 & 65.26 & 62.22 & 60.52 & 60.05\\
    YOLO-v2 \cite{redmon_2017_yolo_CVPR} & 98.05 & 97.55 & 94.94 & 85.93 & 77.84 & 75.37 & 73.40 & 70.84 & 68.42 & 67.20 & 66.91\\
    DeViSE \cite{frome_2013_NIPS} & 97.62 & 97.11 & 94.36 & 84.28 & 75.55 & 73.07 & 70.80 & 67.74 & 64.97 & 63.47 & 63.18\\
    HXE $\alpha$=0.1 \cite{bertinetto_2020_CVPR} & 98.16 & 97.74 & 95.32 & 86.99 & 79.70 & \cellcolor{highlight}77.61 & \cellcolor{highlight}75.69 & 73.22 & 70.96 & 69.78 & 69.49\\
    HXE $\alpha$=0.6 \cite{bertinetto_2020_CVPR} & \cellcolor{highlight}98.32 & \cellcolor{highlight}97.91 & \cellcolor{highlight}95.44 & 86.44 & 78.43 & 75.92 & 73.66 & 70.55 & 67.57 & 65.83 & 65.46 \\
    Soft-labels $\beta$=30 \cite{bertinetto_2020_CVPR} & 98.05 & 97.59 & 95.12 & 86.88 & \cellcolor{highlight}79.76 & 77.54 & 75.62 & 73.25 & 70.95 & 69.75 & 69.45\\
    Soft-labels $\beta$=4 \cite{bertinetto_2020_CVPR} & 98.01 & 97.53 & 94.95 & 84.85 & 75.78 & 73.15 & 70.69 & 67.52 & 64.17 & 61.56 & 61.01\\
    Chang et al. \cite{flamingo_2021_Chang} & 98.10 & 97.55 & 95.01 & 85.87 & 78.04 & 75.45 & 73.34 & 70.82 & 68.20 & 66.76 & 66.43\\
    \ours & \cellcolor{highlight}98.11 & 97.61 & 95.12 & \cellcolor{highlight}87.01 & 79.64 & 77.42 & 75.55 & \cellcolor{highlight}73.26 & \cellcolor{highlight}71.07 & \cellcolor{highlight}69.80 & \cellcolor{highlight}69.53 \\
    \hline
    Cross-Entropy & ~~98.21~~ & ~~97.82~~ & ~~95.32~~ & ~~87.03~~ & ~~79.68~~ & ~~77.42~~ & ~~75.49~~ & ~~73.10~~ & ~~70.79~~ & ~~69.52~~ & 69.24\\
    YOLO-v2 \cite{redmon_2017_yolo_CVPR} & 98.12 & 97.64 & 94.97 & 85.90 & 77.36 & 74.89 & 72.84 & 70.23 & 67.78 & 66.53 & 66.23\\
    HXE $\alpha$=0.1 \cite{bertinetto_2020_CVPR} & 98.25 & 97.87 & 95.46 & 87.07 & 79.65 & 77.51 & 75.55 & 73.14 & 70.85 & 69.65 & 69.34\\
    HXE $\alpha$=0.6 \cite{bertinetto_2020_CVPR} & \cellcolor{highlight}98.39 & \cellcolor{highlight}97.98 & \cellcolor{highlight}95.53 & 86.50 &  78.43 & 75.92 & 73.62 & 70.54 & 67.44 & 65.62 & 65.25 \\
    Soft-labels $\beta$=30 \cite{bertinetto_2020_CVPR} & 98.19 & 97.78 & 95.24 & 87.05 & \cellcolor{highlight}79.78 & \cellcolor{highlight}77.53 & \cellcolor{highlight}75.62 & \cellcolor{highlight}73.30 & 70.94 & 69.71 & 69.40\\
    Soft-labels $\beta$=4 \cite{bertinetto_2020_CVPR} & 94.57 & 93.09 & 80.44 & 62.97 & 42.39 & 33.32 & 30.49 & 25.48 & 21.56 & 17.43 & 17.09\\
    Chang et al. \cite{flamingo_2021_Chang} & 98.24 & 97.77 & 95.22 &  85.97 & 77.98 & 75.44 & 73.35 & 70.78 & 68.12 & 66.64 & 66.30\\
    \ours & 98.25 & 97.76 & 95.28 & \cellcolor{highlight}87.12 & 79.65 & 77.37 & 75.48 & 73.24 & \cellcolor{highlight}70.97 & \cellcolor{highlight}69.74 & \cellcolor{highlight}69.43\\
    \hline
    \end{tabular}
}
\end{center}
\caption{Coarse-classification accuracy results on the test set of \textit{tieredImageNet-H}. The \textit{Top} block reports results without using CRM \cite{karthik_2021_ICLR} and the \textit{Bottom} block reports results using CRM on all coarse-levels from level-1(Coarse (L1)) to and level-11(Fine (L11)).}
\label{tab:tiered_coarse}
\end{table}

\bibliographystyle{splncs04}
\bibliography{bibliography}
\end{document}